\documentclass[twocolumn, switch]{article} 

\usepackage{preprint}

\usepackage[utf8]{inputenc}	
\usepackage[T1]{fontenc}	
\usepackage{amsmath, amsthm, amssymb, amsfonts, mathrsfs}
\usepackage{algorithm}
\usepackage{algorithmicx}
\usepackage{algpseudocode}
\usepackage{xcolor}		
\usepackage{booktabs} 		
\usepackage{nicefrac}		
\usepackage{microtype}		
\usepackage[switch]{lineno}	
\usepackage[colorlinks = true,
linkcolor = purple,
urlcolor  = blue,
citecolor = cyan,
anchorcolor = black]{hyperref}	
\usepackage{array}
\usepackage{multirow}
\usepackage{textcomp}
\usepackage{stfloats}
\usepackage{url}
\usepackage{verbatim}
\usepackage{graphicx}
\usepackage{graphics}
\usepackage{subcaption}
\usepackage{xcolor}
\usepackage{soul}
\usepackage{tikz,hyperref}

\usepackage[numbers,square]{natbib}
\usepackage{titlesec}
\titlespacing\section{0pt}{12pt plus 3pt minus 3pt}{1pt plus 1pt minus 1pt}
\titlespacing\subsection{0pt}{10pt plus 3pt minus 3pt}{1pt plus 1pt minus 1pt}
\titlespacing\subsubsection{0pt}{8pt plus 3pt minus 3pt}{1pt plus 1pt minus 1pt}

\definecolor{lime}{HTML}{A6CE39}
\DeclareRobustCommand{\orcidicon}{
	\begin{tikzpicture}
	\draw[lime, fill=lime] (0,0) 
	circle [radius=0.16] 
	node[white] {{\fontfamily{qag}\selectfont \tiny ID}};
	\draw[white, fill=white] (-0.0625,0.095) 
	circle [radius=0.007];
	\end{tikzpicture}
	\hspace{-2mm}
}
\foreach \x in {A, ..., Z}{\expandafter\xdef\csname orcid\x\endcsname{\noexpand\href{https://orcid.org/\csname orcidauthor\x\endcsname}
			{\noexpand\orcidicon}}
}

\title{Motion Planning of Cooperative Nonholonomic Mobile Manipulators}

\usepackage{xwatermark}
\newwatermark[firstpage,color=gray!90,angle=0,scale=0.28, xpos=0in,ypos=-5in]{*correspondence: \texttt{keshabpatra19@gmail.com}}

\usepackage{authblk}

\author[1\thanks{\tt{keshabpatra19@gmail.com}}]{Keshab Patra\orcidA{}}
\author[2]{Arpita Sinha\orcidB{}}
\author[1]{Anirban Guha}

\affil[1]{Department of Mechanical Engineering,
	Indian Institute of Technology Bombay,
	Mumbai, Maharashtra, India}
\affil[2]{Center for Systems and Control,
	Indian Institute of Technology Bombay,
	Mumbai, Maharashtra, India}

\begin{document}

\twocolumn[ 
  \begin{@twocolumnfalse} 
  
\maketitle

\begin{abstract}
We propose a real-time implementable motion planning framework for cooperative object transportation by nonholonomic mobile manipulator robots (MMRs) in dynamic environments. Our global planner finds a path from start to goal through the static, obstacle-free regions in the environment and generates a set of convex, static, obstacle-free regions around the path using a novel, fast, and computationally lightweight ellipse-based technique. We introduce a nonlinear Model Predictive Control (NMPC) based real-time implementable planning technique that jointly plans feasible motion for the mobile base and the manipulator's arm and generates a kinodynamic feasible, collision-free trajectory for cooperative object transportation. Simulation and hardware experiments validate the efficiency of our proposed planning framework. 
\end{abstract}
\vspace{0.35cm}

  \end{@twocolumnfalse} 
] 


\section{Introduction}
Robotic systems became integral to automation in manufacturing, remote exploration, warehouse management, and other areas. Cooperative multiple MMRs garner attention due to their low cost, small size, redundancy in heavy or oversized object transportation, and fixture-less multipart assembly requiring more Degrees of Freedom (DoF). A cooperative MMR system extends workspace coverage, flexibility, and redundancy with added complexity in robot coordination, communication, and motion planning. Multiple MMRs leverage the mobile bases' locomotion ability and the arms' manipulation ability for object transportation and manipulation in a large workspace.

Nonholonomic mobile bases are widespread in robotic applications because of their advantages in a reduced number of actuators, simplified wheels, and better load-carrying capacity. Nonholonomic MMRs can work better than their holonomic counterparts on uneven ground surfaces as they restrict sideways motion, which leads to increased stability, traction, and controllability on uneven surfaces. The nonholonomic mobile base of the MMRs restricts sideways motion, including non-integrable kinematic constraints. Hence, there are more intricacies in cooperative motion planning and trajectory generation than in the holonomic counterpart.

The study on collaborative manipulators started with a virtual linkage model \cite{1996_khatib} representing the collaborative manipulation systems to generate closed-chain constraints \cite{2023_Xu} between the object and the MMRs for motion synchronization and coordination. The dual arm cooperative control problem has been addressed by NMPC \cite{2024_Zhao}. The coordination scheme for multi-MMR cooperative manipulation and transportation comprises of centralized \cite{2013_Erhart}, decentralized \cite{2018_Culbertson,2018_Verginis} and distributed \cite{2017_Dai,2018_Marino,2020_Ren} control framework. Task allocation \cite{2024a_Keshab,2025_Du} algorithm ensures efficient utilization of the capabilities of the cooperative manipulators. The collision-free navigation started with a variational-based method \cite{1997_Desai} that demonstrated static obstacle avoidance for a two MMR system with poor scalability. Dipolar inverse Lyapunov functions fused with the potential field-based navigation function \cite{2003_Tanner} plan collision-free motion in static environments to transport deformable material by multiple MMRs with a little scope of formation control.

Optimization-based motion planning technique \cite{2017_AlonsoMora} for holonomic MMRs in dynamic environments uses obstacle-free convex polygons around the formation in the position-time space. It optimizes the holonomic MMRs' pose to retain the cooperative MMR system inside the obstacle-free polygon. A geometric path planning approach \cite{2018_Cao, 2017_Jiao} for multiple MMRs transport an object avoiding static obstacles. A rectangular passageway-based approach \cite{2018_Cao} finds the optimal system width and moving direction in the static obstacle-free area for navigation. These methods do not include motion constraints to provide guaranteed feasible motion for nonholonomic MMRs.

A kinematic motion planning technique \cite{2019_Tallamraju} plans for spatial collaborative payload manipulation using a hierarchical approach. The technique's conservative approximation of the obstacles as uniform cylinders highly restricts navigation in tight spaces with high aspect ratios polygonal obstacles. MPC-based motion planning techniques for static obstacle avoidance have been presented in \cite{2017_Nikou,2024_Kennel}. An alternating direction method of multipliers-based distributed trajectory planning algorithm \cite{2020_Shorinwa} plans trajectory in a static environment. A distributed formation control technique \cite{2021_Wu} utilizes constrained optimization for object transportation in a static environment. Motion Planning for deformable object transportation \cite{2022_Hu,2024_Pei} in a static environment uses optimization techniques. A reciprocal collision avoidance algorithm \cite{2020_Mao} combined with MPC doesn't maintain any formation. These generic planning algorithms cannot be used as they do not maintain the rigid formation required for collaborative MMRs. An NMPC-based kinodynamic motion planning technique \cite{2024b_Keshab} plans motion for object transportation by multiple MMRs in dynamic environments. The proposed planning technique is limited to holonomic MMRs and environments with convex static obstacles.

We propose an end-to-end trajectory planning framework for collaborative object transportation by nonholonomic MMRs. Our proposed algorithm plans the trajectory in two steps: offline path planning and online motion planning. To the best of our knowledge, the end-to-end planning for nonholonomic MMRs has not been done before. The offline planning algorithm computes a linear, piecewise path from start to goal using the visibility vertices algorithm \cite{2005_Choset} and defines the static, obstacle-free region for motion planning optimization. Motivated by \cite{2014_Deits}, we have developed a novel, fast, ellipse-based optimization-free convex polygon computation algorithm to define the static obstacle-free region around the piecewise linear path. Starting from computing a polygon around the path segments using its visibility vertices, we convexify it by eliminating the concave vertices using the tangents of ellipses. We compute the ellipse, aligning its major axis along the path segment that touches the nearest concave vertex, and inflate it to eliminate subsequent concave vertices in the same polygon. The proposed planning technique eliminates the restriction of convex obstacles of IRIS-based planning algorithms \cite{2014_Deits, 2017_AlonsoMora, 2024b_Keshab} and convex optimization, guaranteeing the path segment will remain within it.

The major contributions of this article are as follows.

\begin{enumerate}
	\item We propose a fast ellipse-based optimization-free convex polygon computation algorithm to define the static obstacle-free region around the linear piece-wise path using its visible vertices. The convex obstacle-free region is regarded as inequality constraints in the NMPC for motion planning.
	\item We introduce an NMPC-based, real-time implementable online motion planner that jointly plans for the nonholonomic MMR's base and the manipulator. The planner computes a kinodynamic feasible collision-free motion plan for the multiple MMRs in a dynamic environment.
\end{enumerate}
\section{Problem Formulation}\label{sec:ProblemFormulation}
\begin{figure}[htbp]
	\centerline{\includegraphics[width = 220px]{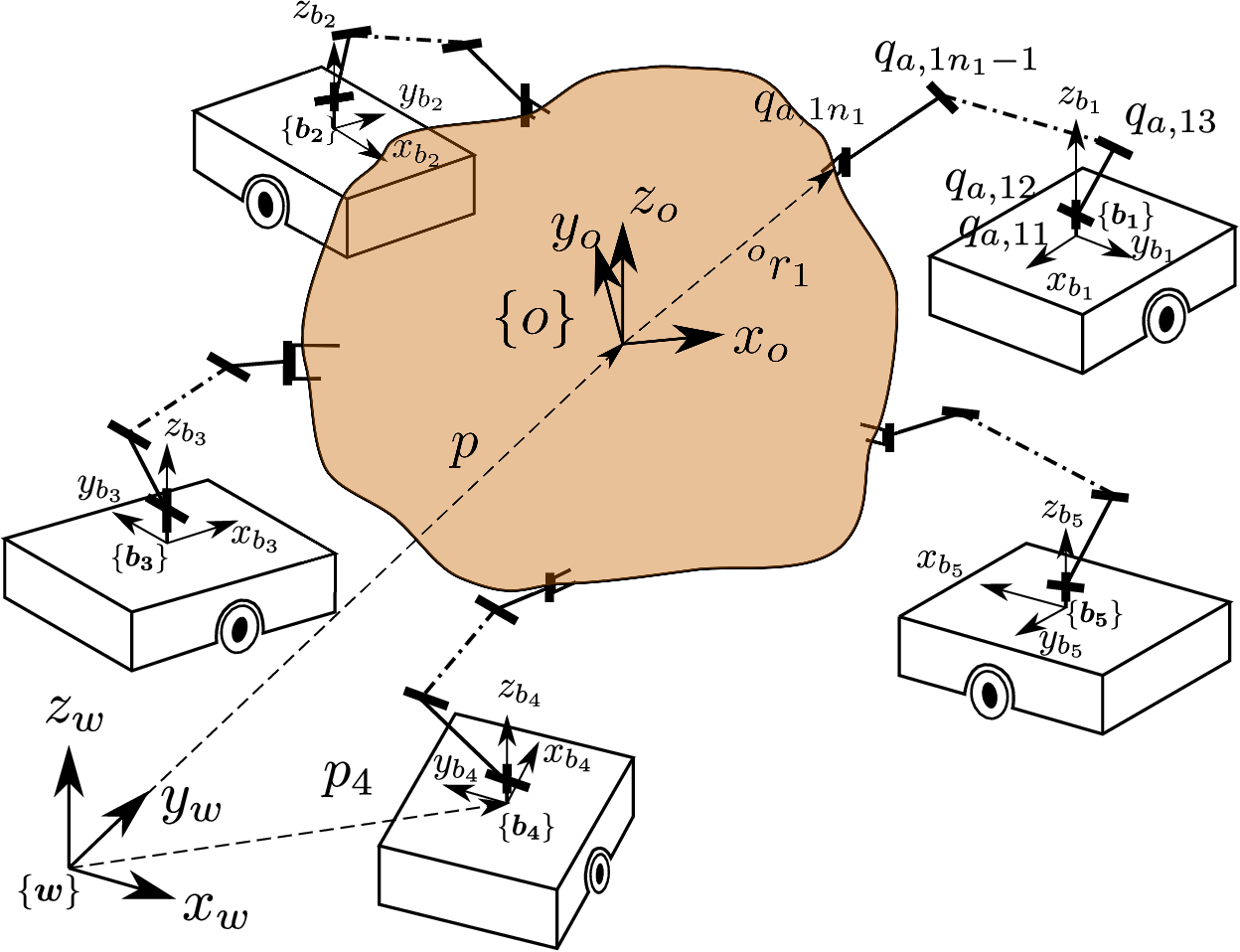}}
	\caption{Formation of five non-holonomic MMRs holding an object. The MMRs grasped the object to transport collaboratively without any collision.}
	\label{fig:1}
\end{figure}

A system of $n$ nonholonomic MMRs grasps a rigid object at its periphery as shown in Fig. \ref{fig:1} to collaboratively transport the object without any collision.

$\{\boldsymbol{w}\}$ defines the world fixed reference frame. An object coordinate frame $\{\boldsymbol{o}\}$ is attached to the object center of mass (CoM), and each MMR has its own body coordinates $\{\boldsymbol{b}_i\}$ attached to the center of its mobile base. Without specific mention, all the quantities are defined in $\boldsymbol{\{w\}}$. The collaborative manipulation system is defined in the following subsections.

\subsection{Mobile Manipulator and Collaborative Formation}
The mobile base of $i$-th MMR is defined with pose $q_{m,i}=[p^T_i,\phi_i]^T$ where $p_i\in\mathbb{R}^2$ and $\phi_i\in\mathbb{R}$ are the position and orientation of the mobile base. The manipulator of $i$-th MMR has $n_i$ number of joints and it's displacement is defined as $q_{a,i} = [q_{a,i1},q_{a,i2},\cdots, q_{a,in_i}]^T$. The whole MMR is defined by $q_i=[q_{m,i}^T,q_{a,i}^T]^T$. The $i$-th EE's position and orientation is defined in $\{w\}$ as $p_{ee,i}\in\mathbb{R}^3$ and $\phi_{ee,i}\in\mathbb{R}^3$. 
The $i$-th non-holonomic MMR's first-order dynamics is $\dot{q}_i =[v_{i}\cos(\phi_i), v_{i}\sin(\phi_i),\omega_{i},\dot{q}_{a,i}]^T$ where the control inputs are mobile base's linear and angular velocities $v_{i}$, $\omega_{i}$ respectively and the manipulator's joint velocities $\dot{q}_{a,i}$, therefore, $u_i = [v_{i},\omega_{i},\dot{q}_{a,i} ]$. 

We represent the coupled first order system dynamics for $i$-th MMR  by a discrete-time non-linear system as
\begin{equation}\label{stf}
	q^{k+1}_i =f(q^k_i ,u^k_i)
\end{equation}
where $k$ is the discrete time step. The admissible states and control inputs are defined by Eqn. \ref{eqn:limits}

\begin{equation}\label{eqn:limits}
	\underline{q}_{a,i}\leq q_{a,i}\leq\overline{q}_{a,i},\  
	\underline{u}_{i}\leq u_{i}\leq\overline{u}_{i},\ \forall i \in [1,n]
\end{equation}

where $\underline{q}_{a,i}$, $\overline{q}_{a,i}$ represents the manipulator's joint position limit vector and $\underline{u}_{i}$, $\overline{u}_{i}$ are the admissible control limits. The set of admissible states $\mathcal{Q}_i$ and control inputs $\mathcal{U}_i$ are indicated by joint position and velocity vectors' limit (Eqn. \eqref{eqn:limits}), $\mathcal{Q}_i = [\underline{q}_{a,i}, \overline{q}_{a,i}]$, $\mathcal{U}_i = [\underline{u}_{i}, \overline{u}_{i}]$

The EE of the $i$-th MMR of the multi-MMR formation $\mathcal{F}$ (Fig \ref{fig:1}) grasps the object at $^or_i$ defined in object frame $\boldsymbol{o}$, where the superscript $\boldsymbol{o}$ indicates it's reference frame $\{\boldsymbol{o}\}$. The formation configuration is defined by $\mathcal{X}=[p^T,o^T,Q^T]^T$, where $p\in\mathbb{R}^3$ is the position and $o\in\mathbb{R}^3$ is the orientation of the object CoM, $Q=[q_1^T,q_2^T,\cdots, q_n^T]^T$ is the configuration of $n$ MMRs. The space occupied by the formation is defined as $\mathcal{B}(\mathcal{X})$.

\subsection{Environments}
A structured and bounded environment having both static and dynamic obstacles is defined as $W$. $\mathcal{O}$ represents the set of static obstacles in $W$. The static obstacle-free workspace is defined by $    W_{free} = W\setminus\mathcal{O}\in\mathbb{R}^2$.
The set of dynamic obstacles in the environment is defined as $\mathcal{O}_{dyn}$. The start position $p_s$ and the goal position $p_g$ of the object CoM are in the obstacle-free space $W_{free}$.

The planning objective is to design a motion planning
framework such that
1) the MMRs can cooperatively transport the object  without any collision.
2) the generated trajectory is kinodynamically feasible and
within the admissible limits of the MMRs minimizing the control input of MMRs.
3) the planner can handle any rigid object, grasping configuration and static concave obstacles directly.
\section{Motion Planning}\label{sec:MotionPlanning}
\begin{figure}[htbp]
	\centerline{\includegraphics[width = 240px]{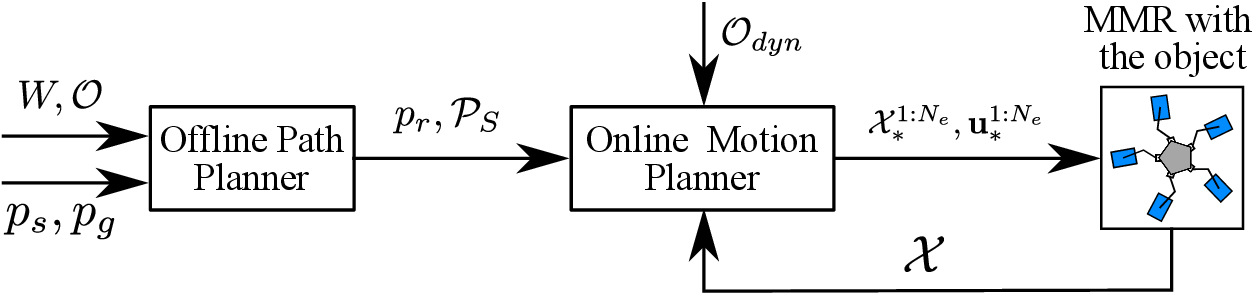}}
	\caption{Two step motion planning process: offline path planning and online motion planning.}
	\label{fig2}
\end{figure}

We solve the motion planning problem for cooperative multi-MMRs in two steps: offline path planning and online motion planning shown in Fig. \ref{fig2}. In the offline path planning step, we compute a static obstacle-free linear piece-wise shortest distance path ($S$) between the start and the goal location using offline path planner in Section \ref{GPP}. Then we compute a set of connected convex region around the path using our proposed convex polygon computation algorithm in \ref{sec:CPC}.

In the next step, an online motion planner (Section \ref{OMP}) computes a feasible motion plan for the collaborative MMRs in receding horizons. The planner generates a kinodynamically feasible trajectory in the dynamic environment using $p_r(c_t)$ as an initial guess. The generated trajectory is free from collision with the static and dynamic obstacles and collision among the MMRs and with the object.

\begin{figure}[htbp]
	\centerline{\includegraphics[width = 230px]{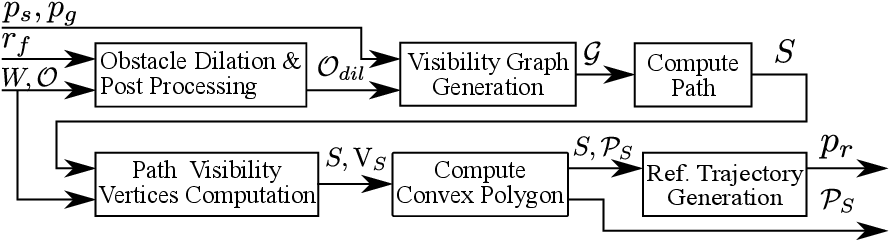}}
	\caption{Offline path planning and convex polygon computation process around $S$.}
	\label{fig:3}
\end{figure}

\subsection{Global Path Planner}\label{GPP}
The global path planner computes a static obstacle-free path $S$ for the MMRs' between the start and goal in offline employing visibility vertices finding algorithm \cite{2005_Choset}. Then it computes a set of connected convex obstacle-free polygon around $S$. Fig. \ref{fig:3} shows the outline of the path planning process.

\subsubsection{Path Computation} 
\begin{figure}[H]
	\centerline{\includegraphics[width = 220px]{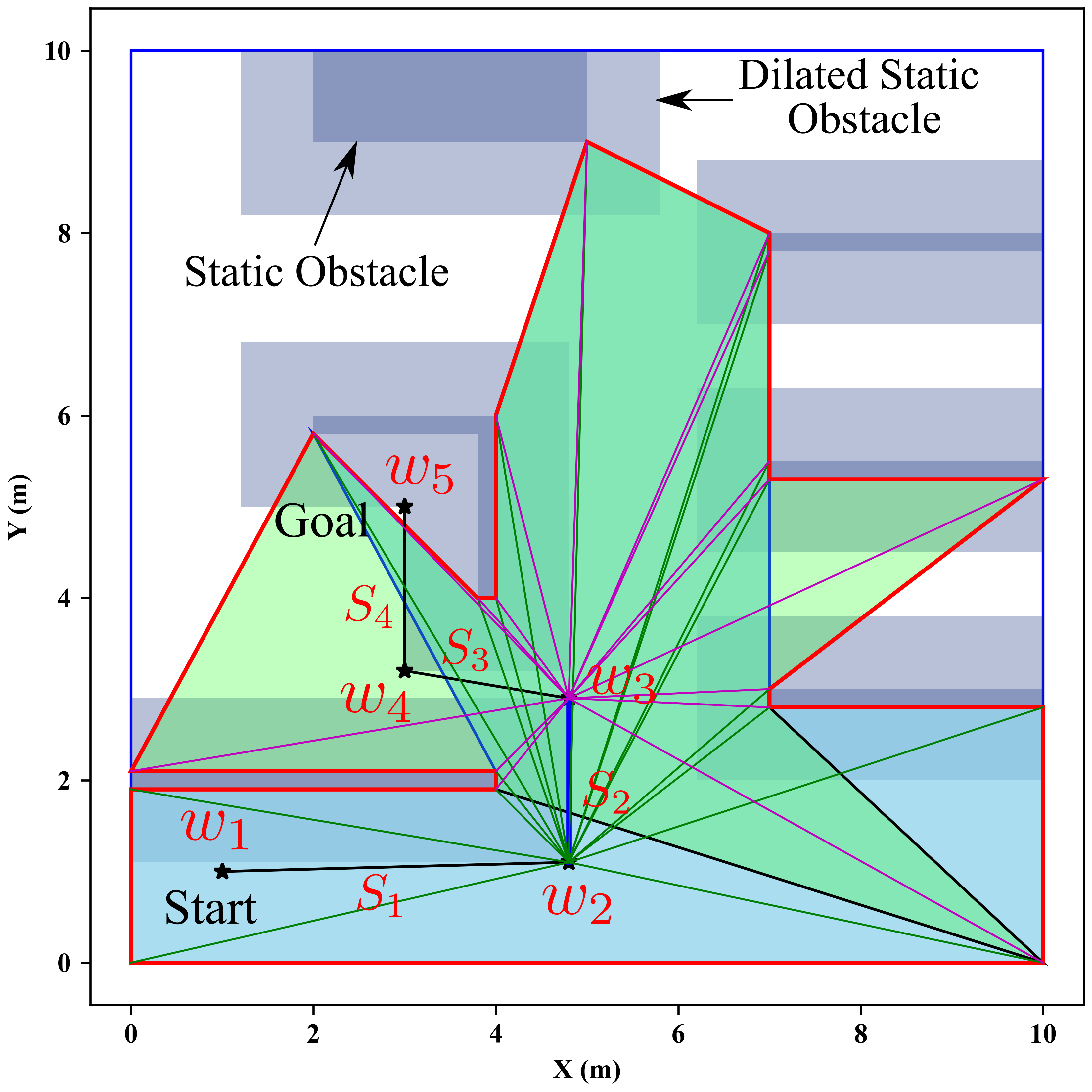}}
	\caption{Path Polygon for $S_2$ computed using the visible vertices of $W_2$ and $W_3$.}
	\label{fig:4}
\end{figure}

The global path planner dilates $\mathcal{O}$ by a distance $r_f$ so that we consider the formation $\mathcal{F}$ at any point $p$,the CoM of the grasped object. The dilation distance $r_f$ is the radius of a circle located at $p$, inside which the formation could always be enclosed. The planner substitutes the mutually intersecting dilated obstacles $\mathcal{O}_{dil}$ with their union. A visibility vertices finding algorithm \cite{2005_Choset} creates a visibility map considering $\mathcal{O}_{dil}$ and the start and goal point. The vertices from the visibility map are added as node $\mathcal{V}$ to a graph $\mathcal{G}(\mathcal{V}, \mathcal{E}, \mathcal{W})$. An edge $\mathcal{E}$ is added between two mutually visible nodes using the visibility map with the Euclidean distance between them as weight $\mathcal{W}$. A graph search algorithm computes the shortest linear piece-wise path $S$ between the start $p_s$ and the goal $p_g$ location. Fig. \ref{fig:4} shows the computed path $S$ with linear segments $S_1, S_2, S_3, S_4$ and vertices $w_1, w_2, w_3, w_4, w_5$.

\subsubsection{Convex Polygon Computation}\label{sec:CPC}
We compute a static obstacle-free polygon around a path segments $S_i\in S$ and convexify. We consider $\mathcal{O}$ and the vertices $w_i$ of the path to obtain the set of vertices $\mathrm{V}_s$ present in the static environment and visible from $w_i,\ \forall i$ (Fig. \ref{fig:4}). A simple polygon is defined for each vertex $w_i,\ \forall i$ by cyclically connecting its visible vertices. The polygon remains in  $W_{free}$. The union of polygons obtained for $w_i$ and $w_{i+1}$ of a path segment $S_i$ defines a static obstacle-free simple polygon $\mathrm{P}_{cc,i}$ around $S_i\in S$. Fig. \ref{fig:4} shows the computed polygons for the path segment $S_2$ using the polygons around $w_2$ and $w_3$. The union of the two polygons (green and blue) in red boundary defined around $w_2$ and $w_3$ provides a static obstacle-free polygon $\mathrm{P}_{cc,2}$ around $S_2$. A set of polygon $\mathcal{P}_{cc}$ for $S$ is computed similarly. The computed polygons are generally concave. Convexification is needed for the static obstacle avoidance constraints in motion planning optimizations. We compute a set of convex polygons $\mathcal{P}_S$ analytically as a subset of their original concave polygons in $\mathcal{P}_{cc}$. To eliminate the concave vertices of $\mathrm{P}_{cc,i}$ we use tangents of ellipses at concave vertices, whose major axis aligned with $S_i$. The convexification steps are illustrated in the Algo. \ref{algo:convexify}.
 
We define an ellipse in the ground plane as
\begin{equation}
	\kappa(\mathrm{C},\mathrm{d}) = \{x = \mathrm{C} \overline{x} + \mathrm{d}:\ ||\overline{x}||\leq 1, x\in \mathbb{R}^2\}
\end{equation}
where $\mathrm{C}$ is a $2\times2$ symmetric positive definite matrix that maps the deformation of a unit radius circle ($||\overline{x}||\leq 1$) to an ellipse. $\mathrm{C}$ is decomposed as $\mathrm{C} = R^T\Lambda R$, where $R$ is a rotation matrix that aligns the ellipse axes to the world reference frame axes and $\Lambda = diag(a,b)$ is a diagonal scale matrix. The diagonal elements $a$ and $b$ of $\Lambda$ refer to the length of the ellipse semi-major and minor axes. $\mathrm{d}$ defines the center of the ellipse.

\begin{algorithm}
	\caption{Polygon Convexification}\label{algo:convexify}
	\begin{algorithmic}[1]
		\Statex \textbf{Input:} Concave Polygon $\mathrm{P}_{cc,i}$, Path segment $S_i$
		\Statex \textbf{Output:} Convex Polygon $\mathrm{P}$
		\Statex\hrulefill
		\State $\mathrm{P}\gets\mathrm{P}_{cc,i};\ j\gets 0$
		\State $\mathcal{V}_{cc}\gets$ ConcaveVertices($\mathrm{P}$)
		\If{$\mathcal{V}_{cc}\neq\phi$}
		\State $\mathrm{d}\gets$ Midpoint($S_i$); $R$: major axis along $S_i$
		\State $a\gets 0.5length(S_i) + r_f$
		\State $x^*_j \gets$ NearestVertex($\mathcal{V}_{cc}, \mathrm{d}$)
		\State $\mathrm{C}_j, d\gets$ FindEllipse($R,a,\mathrm{d},x^*_j$) \Comment{Use $||\overline{x}^*_j||=1$}
		\State $\kappa_j\gets (\mathrm{C}_j,\mathrm{d})$
		\State $ \mathrm{a}_j\gets2\mathrm{C}^{-T}_j\mathrm{C}^{-1}_j(x^*_j-\mathrm{d}); b_j\gets \mathrm{a}^T_jx^*_j$
		\Comment{Tangent of ellipse $\kappa_j$ at $x^*_j$}
		\State $\mathrm{P},\mathcal{V}_{cc}\gets$ DiscardVertices$(\mathrm{a}_j,b_j, \mathrm{P},\mathcal{V}_{cc})$
		\While{$\mathcal{V}_{cc}\neq\phi$}
		\State $j = j+1$
		\State $x^*_j=$ NearestVertex($\mathcal{V}_{cc}, \mathrm{d}$)
		\State $\kappa_j\gets$ DilateEllipse($\kappa_0, R, x^*_j$)
		\State repeat line $9-10$
		\EndWhile
		\State $\mathrm{P}:\mathbf{A}\gets[\mathrm{a}^T_0,\mathrm{a}^T_1, \cdots]^T,\mathbf{b}\gets[b_0,b_1,\cdots]^T$
		\Else
		\State  $\mathrm{P}:(\mathbf{A},\mathbf{b})\gets$ HalfPlanes($\mathrm{P}$)
		\EndIf
		\State \textbf{return} $\mathrm{P}(\mathbf{A},\mathbf{b})$
		
	\end{algorithmic}
\end{algorithm}

\begin{figure}[H]
	\centering
	\begin{subfigure}[t]{0.24\textwidth}
		\includegraphics[width=115px]{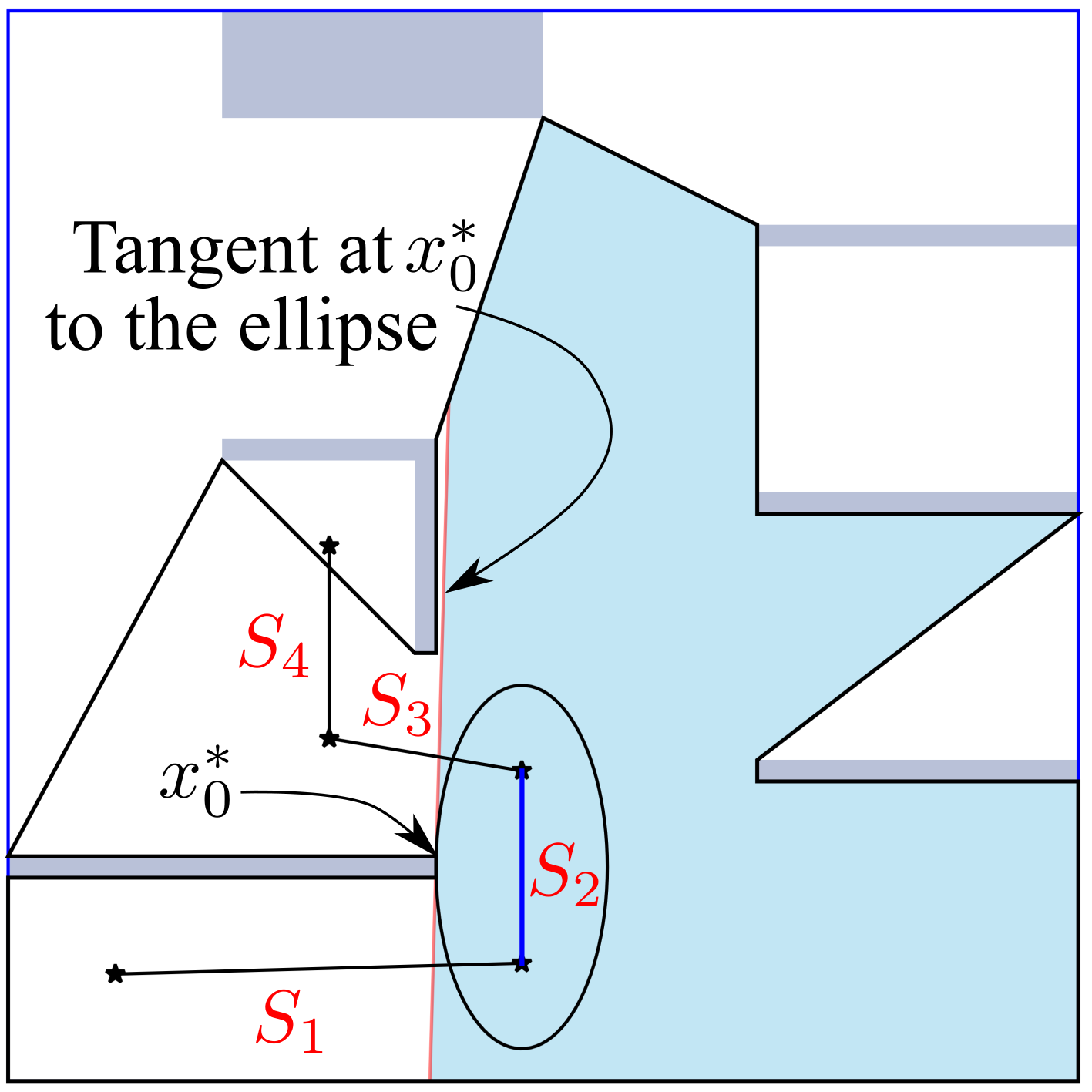}
		\subcaption{\label{fig:5a}}
	\end{subfigure}
	\begin{subfigure}[t]{0.24\textwidth}
		\includegraphics[width=115px]{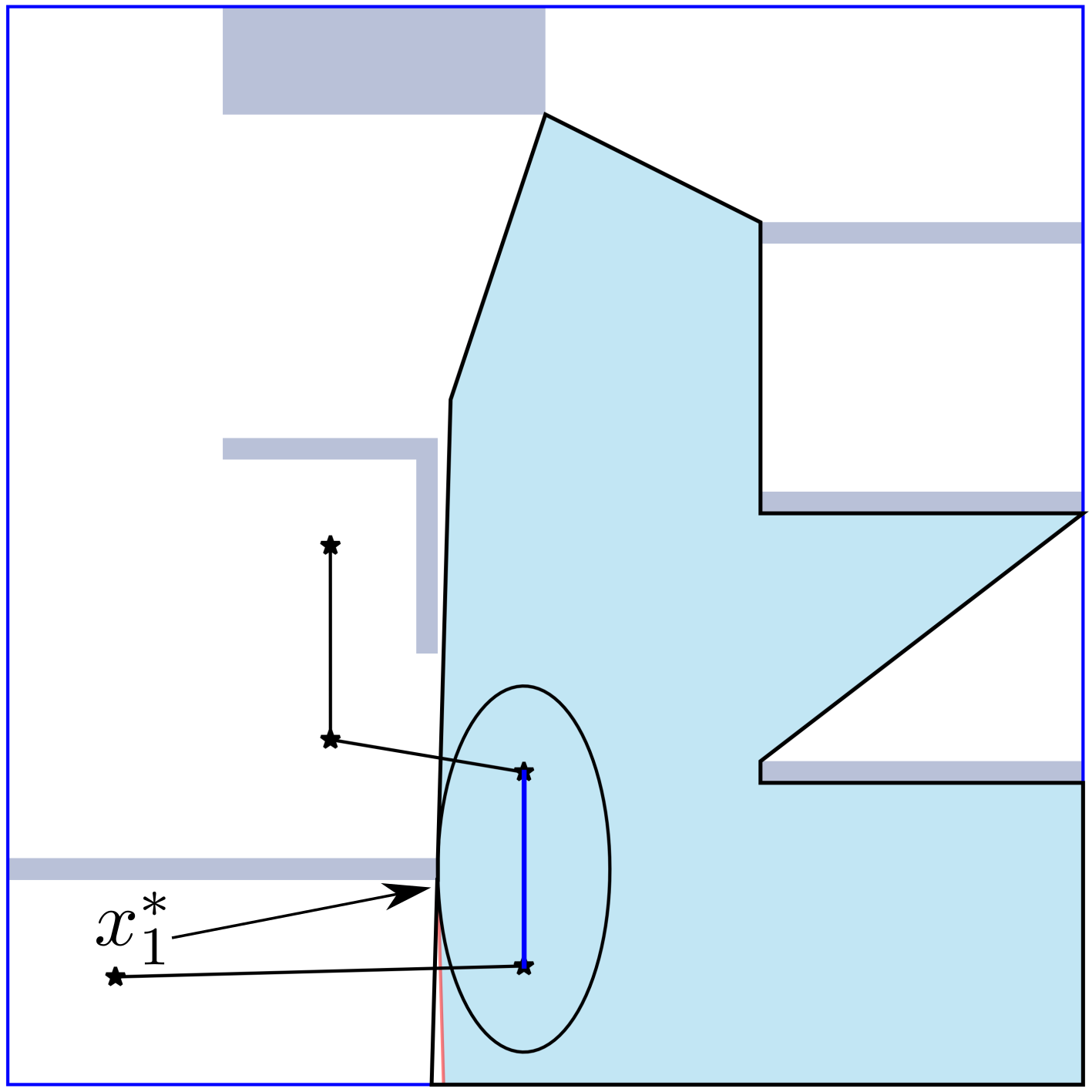}
		\subcaption{\label{fig:5b}}
	\end{subfigure}
	\begin{subfigure}[t]{0.24\textwidth}
		\includegraphics[width=115px]{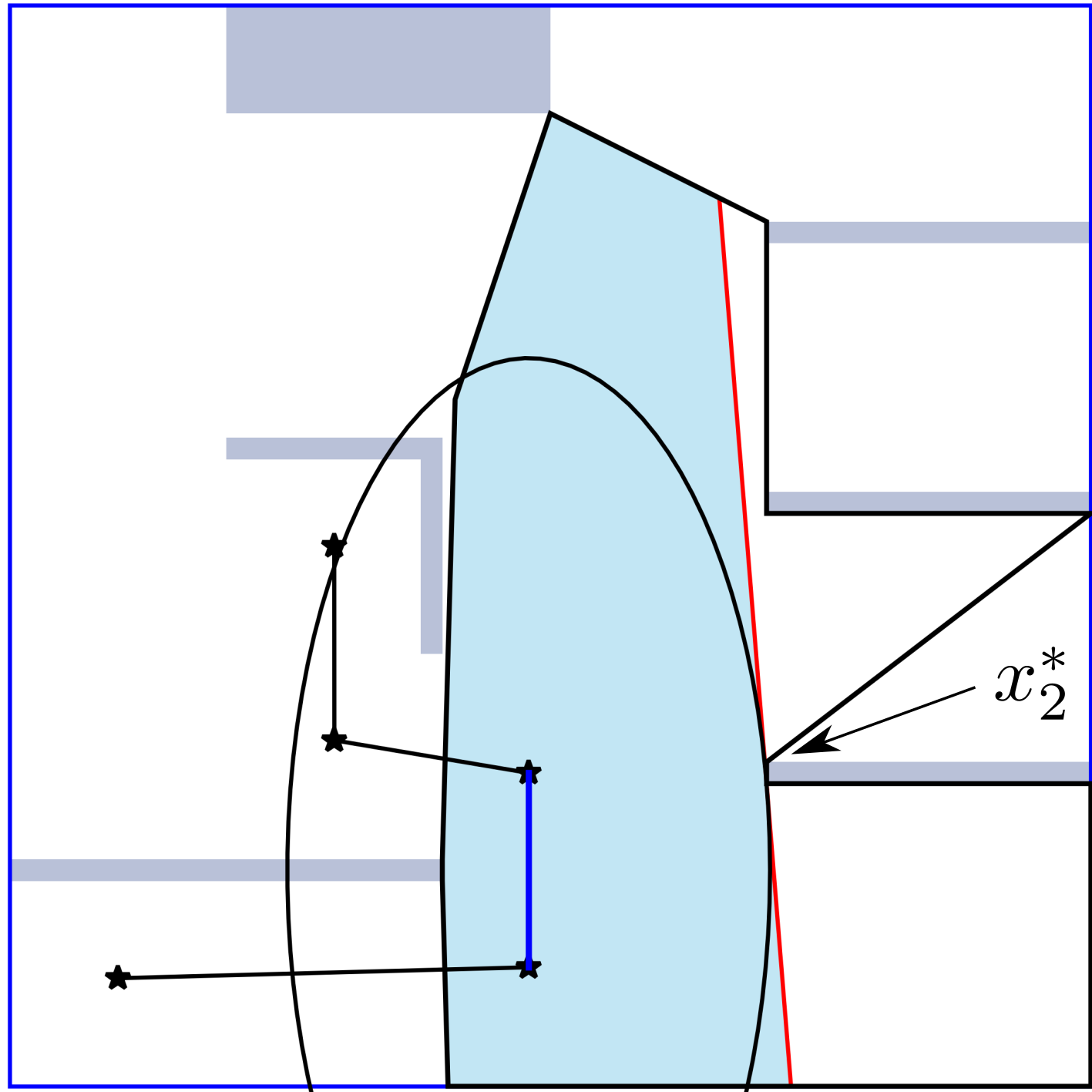}
		\subcaption{\label{fig:5c}}
	\end{subfigure}
	\begin{subfigure}[t]{0.24\textwidth}
		\includegraphics[width=115px]{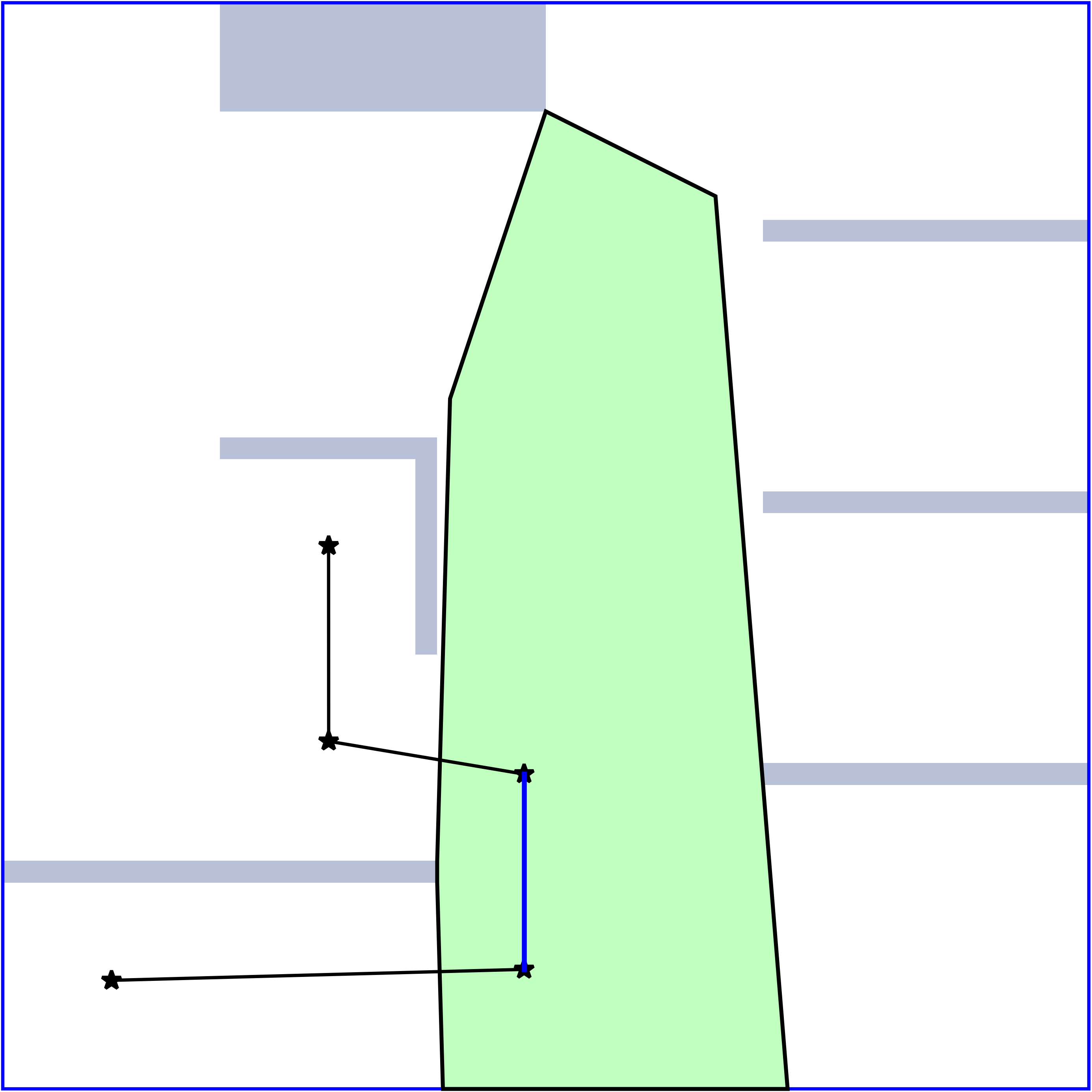}
		\subcaption{\label{fig:5d}}
	\end{subfigure}
	\begin{subfigure}[t]{0.48\textwidth}
		\includegraphics[width = 200px]{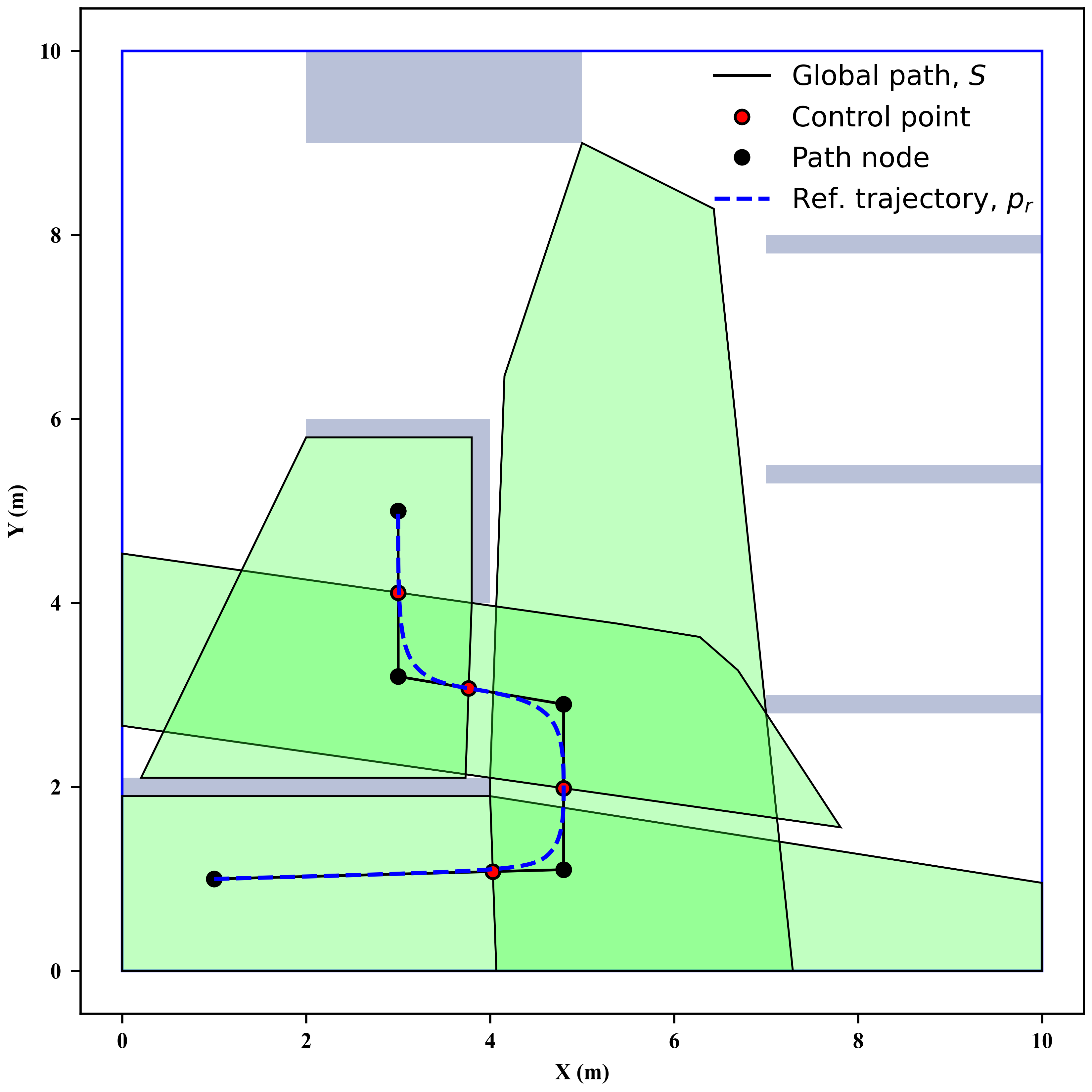}
		\subcaption{\label{fig:5e} Convex polygons (light green) around $S$.}
		
	\end{subfigure}
	\caption{Fig. \ref{fig:5a} - \ref{fig:5d} shows the steps of polygon convexification process for $S_2$. A tangent of an ellipse touching the nearest concave vertex (red line) of the polygon eliminates its concavity by cutting the polygon (black edges). The polygon (sky blue) containing the path segment has been kept. A convex polygon is formed (green polygon in Fig. \ref{fig:5d}.)}
	\label{fig:5}
\end{figure}

Algo. \ref{algo:convexify} initializes a polygon $\mathrm{P}$ with the concave polygon $\mathrm{P}_{cc,i}$. It finds the concave vertices $\mathcal{V}_{cc}$ of $\mathrm{P}$. If $\mathcal{V}_{cc}$ is not empty, then it convexifies $\mathrm{P}$ in line $4-17$. The algorithm computes the half-plane representation of $\mathrm{P}$ in line $19$. In the polygon convexification process, the algorithm fits an ellipse $\kappa_0$ center $\mathrm{d}$ at the midpoint of the path segment $S_i\in S$. The major axis is aligned with the path segment $S_i$, the semi-major axis length $a = 0.5length(S_i) + r_f$. The ellipse is inflated in the minor-axis direction till it touches the nearest concave points $x^*_0$ to $d$ and an ellipse $\kappa_0$ is computed in line $4-8$. The tangent to the ellipse $\kappa_0$ at point $x^*_0$ defines the inequality $H_0 = \{x: \mathrm{a}^T_0 x \leq b_0\}$. After obtaining $H_0$ in line $9$, we cut the polygon with $H_0$ and keep the polygon that contains $S_i$ and keep it as $\mathrm{P}$ (Fig. \ref{fig:5a}). We discard the concave vertices outside the new polygon $\mathrm{P}$ from $\mathcal{V}_{cc}$. If there is any concave vertices left $\mathcal{V}_{cc}\neq\phi$ the ellipse $\kappa_0$ is dilated to form an ellipse $\kappa_1$ in line $14$ keeping the aspect ratio same till it touches the nearest concave vertices $x^*_i\in\mathcal{V}_{cc}$ to $\mathrm{d}$. The tangent to the ellipse at point $x^*_1$ defines the inequality $H_1 = \{x: \mathrm{a}^T_1 x \leq b_1\}$. After obtaining $H_1$ in line $9$, we cut the polygon with $H_1$ and keep the polygon that contains $S_i$ and keep it as $\mathrm{P}$, as shown in Fig. \ref{fig:5b}. The convexification process in line $11-16$ is repeated until no concave vertex left in the polygon $\mathcal{V}_{cc}\neq\phi$ (Fig. \ref{fig:5b}-\ref{fig:5d}). After eliminating all the concave vertices, the polygon $\mathrm{P}$ becomes convex (Fig. \ref{fig:5d}), and the half-plane representation of $\mathrm{P}$ is returned in line $17$.

The  Fig. \ref{fig:5} shows the polygon convexification process of $\mathrm{P}_{cc,2}$ defined for $S_2$. An ellipse touching its nearest concave vertex to $S_2$, has been obtained and a tangent (red line) to the ellipse at this point has been drawn in Fig. \ref{fig:5a}. The tangent cuts the polygon into two. The polygon (sky blue) containing the path segment has been kept. The ellipse has been dilated keeping the aspect ratio same in Fig \ref{fig:5b} till it touches the nearest concave vertex (to $S_2$) of the new polygon. Here a very small portion of the polygon is cut by the tangent to the ellipse at this concave vertex. The process continues till any concave vertex remains and a convex polygon is formed (green polygon in Fig. \ref{fig:5d}.)

Fig. \ref{fig:5e} shows a set of convex polygons $\mathcal{P}_S$ in light green around the path segment $S$ computed using the Algo. \ref{algo:convexify}. Every segment of $S$ remains inside any convex polygon $\mathrm{P} \in \mathcal{P}_S$ defined in $W_{free}$. We add additional intermediate control points in red dots in Fig. \ref{fig:5e} on $S$. The control points are used to generate time-normalized smooth trajectory $p_r(c_t)$ from the $S$ using a Bézier curve with normalized time parameter $c_t \in [0,1]$. These intermediate control points are inserted when a new convex polygon appears along the path $S$ from the start point. A control point is added for the last path segment while exiting the intersection area of the last two polygons. The generated quadratic Bézier curve would remain in $W_{free}$ as any three consecutive control points remain within a single convex polygon. The time-normalized reference trajectory $p_r(c_t)$ is used as an initial guess for the online motion planner.
  
\subsection{Online Motion Planner} \label{OMP}
The global path planner in Section \ref{GPP} computes a static obstacle-free path $S$. It does not delve into the motion constraints of the MMRs, dynamic obstacle avoidance, and the collision among the MMRs but provide the global references. We propose an online motion planner as a constrained nonlinear optimization problem incorporating kinodynamic constraints. It uses the smoothed global reference trajectory $p_r(c_t)$ as an initial guess to eliminate the local stuck. The online motion planning optimization is given as
\begin{subequations}\label{lo}
        \begin{align}
            \mathcal{X}^{0:N_h}_*,\mathbf{u}^{0:N_h} & =\arg\min \sum_{k=0}^{N_h-1} J( \mathcal{X}^k,u^k) + J_{N_h} \label{lo1}\\
            \textrm{s.t.}\quad q^{k+1} & =f(q^k,u^k) \label{lo2}\\
            \mathcal{B}(\mathcal{X}) & \subset\mathcal{W}_{free} \label{lo3}\\
            \mathcal{B}(\mathcal{X}) & \cap \mathcal{O}_{dyn} = \emptyset  \label{lo4} \\
            H_i & \mathrm{v}_{<x,y>}\leq h_i, \label{lo8}\\
            H_{(i+1)\%n} & \mathrm{v}_{<x,y>} \geq h_{(i+1)\%n}, \label{lo9}\\
            0\leq \mathrm{v}_z & \leq  Z_h ,\forall\mathrm{v}\in\mathscr{V}_i(q_i), \forall i\in [1,n] \label{lo10}\\
            q^k_{a,i} & \in \mathcal{Q}_i , u^k_i \in \mathcal{U}_i, \forall i\in [1,n] \label{lo5}\\
            g_i(\mathcal{X}) & = 0, \forall i\in [1,n]\label{lo6}\\ 
            \mathcal{X}^0 & = \mathcal{X}(0) \label{lo7}
        \end{align}
\end{subequations}
where the superscript $k$ refers to the discrete time step. Section \ref{lcf} illustrates the cost function in Eqn. \eqref{lo1}. Eqn. \eqref{lo2} represents the state transition function (Eqn. \eqref{stf}) of the system. Eqn. \eqref{lo3} and \eqref{lo4} account for the static and dynamic obstacles avoidance constraints detailed in Section \ref{sec:soa} and Section \ref{sec:doa}. The self collision avoidance described in Section \ref{sca} is ensured by the constraints in the Eqn. \eqref{lo8}-\eqref{lo10}. The set of admissible states and control inputs are defined in Eqn. \eqref{lo5} and elaborated in Eqn. \ref{eqn:limits}. The grasp constraints described in Section \ref{gc}) are maintained by Eqn. \eqref{lo6}. Eqn. \eqref{lo7} defines the initial state of the formation in a planning horizon $N_h$.

The online motion planner computes the optimal motion plan for the MMRs by solving the optimization problem in Eqn. \ref{lo} in receding horizons for $N_h$ horizon segment with time $T_h$ to reduce the computational burden. The horizon length should be long enough to capture the sharp turning behavior. The MMRs execute the computed motion plan of a horizon for an execution time $T_e$ $(T_e < T_h)$. We choose the execution time $T_e$ so that the computation time for a horizon is guaranteed to be less than $T_e$, and the MMRs can start the next plan once it finishes executing the current plan. In case of failure to get a motion plan the MMRs would stop motion and the planner would try to re-plan from the stopping position. 
 
\subsubsection{Cost Function} \label{lcf}
The cost function of the optimization in Eqn. \eqref{lo} described in Eqn. \eqref{tcf} minimizes the control inputs and the tracking error with respect to the initial guess trajectory. The diagonal weight-age matrix $\mathbf{W_u}$ for control effort minimization is provided with higher value than the weight-age matrix $\mathbf{W_e}$ to the tracking error ${e^k}$ of the object CoM. The higher weight-age to control inputs prioritize input minimization. The lower weight values to the tracking error provides global guidance to the trajectory with flexibility to deform for dynamic obstacle avoidance and kinodynamic motion compliance.
\begin{equation}\label{tcf}
	J( \mathcal{X}_k,u_k)= {u^k}^T\mathbf{W_u}u^k + {e^k}^T\mathbf{W_e}e^k
\end{equation}

We discretize the reference path $p_r(c_t)$ into $\Lambda$ path segment.   The expected and reference position for the CoM of the object is denoted as $p^k$ and $p^{\lambda+k}_r = [x, y]^{\lambda+k}_r$ for the future time step $k$ and $\lambda$ is the index of the nearest reference path segment to $p^0$. The discretized path should hold the relation $\sum_{k= 0}^{ N_h-1}||p^{\lambda+k+1}_r - p^{\lambda+k}_r||\leq v_{op}T_h$, where $v_{op}$ is the operational velocity of the formation. The tracking error vector $e^k$ is defined as

\begin{equation}
    e^k = p^k - p^{\lambda+k}_r
\end{equation}

The terminal cost $J_{N_h}$ is defined in Eqn. \eqref{tlcf} similar to the tracking error with a higher weighting $\mathbf{W}_{n_h}$.

\begin{equation}\label{tlcf}
    J_{N_h}={e^{N_h}}^T\mathbf{W_{N_h}}e^{N_h}
\end{equation}

\subsubsection{Static Obstacles Avoidance} \label{sec:soa}
\begin{figure}[h]
	\centerline{\includegraphics[width = 220px]{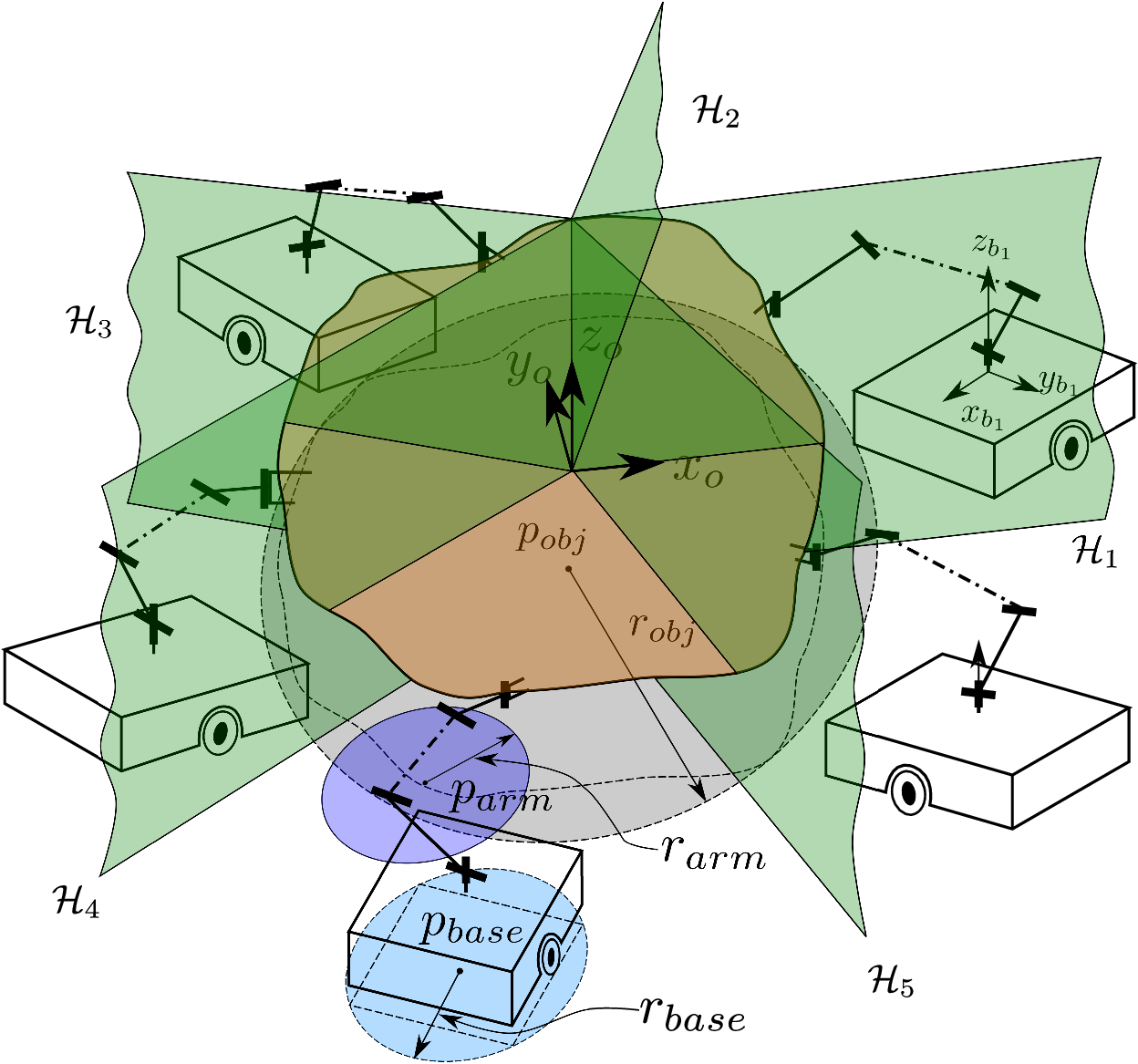}}
	\caption{The infinite convex wedge for $i-th$ MMR is defined by the half plane $\mathcal{H}_i$, $\mathcal{H}_{(i+1)\%n}, z=0$ and $z=\infty$. The enclosing circles for MMRs' mobile base and manipulator are blue and violet, respectively, and the object is gray.}
	\label{fig:soa}
\end{figure}

The formation $\mathcal{F}$ must remain within $W_{free}$ ($\mathcal{B}(\mathcal{X})\subset\mathcal{W}_{free}$) to avoid collision with static obstacles. $W_{free}$ is represented by a set of convex polygons $\mathcal{P}_S$ computed by the offline path planner in Section \ref{sec:CPC}. The inequality representation of the polygon $\mathrm{P} \in \mathcal{P}$ is in Eqn. \eqref{eqn:12}.
\begin{equation}\label{eqn:12}
    \mathrm{P}=\{\textrm{x}\in\mathbb{R}^2:\mathbf{A}\textrm{x}\leq \boldsymbol{b}, \mathbf{A}\in\mathbb{R}^{n_f\times 2}, \boldsymbol{b}\in\mathbb{R}^{n_f}\}
\end{equation}
where $n_f$ is the number of the sides of $\mathrm{P}$ and $\textrm{x}$ is an interior point of $\mathrm{P}$. The set of vertices of the bounding polygons of the object and the $n$ MMRs are represented by $\mathscr{V}(\mathcal{X})$. The projection of $\mathscr{V}(\mathcal{X})$ at the ground plane $(z=0)$ must remain within any polygon $\mathrm{P}\in \mathcal{P}_S$. The constraints are represented as follows

\begin{equation}\label{eqn:soa}
\mathbf{A}\mathrm{v}_{<x,y>}\leq \boldsymbol{b} - d_{safe}, \forall\mathrm{v}\in\mathscr{V}(\mathcal{X}),\mathrm{P}:(\mathbf{A},\boldsymbol{b})
\end{equation}
where $\mathrm{v}_{<x,y>}$ is the $x-y$ projection of the vertex $\mathrm{v}\in\mathscr{V}(\mathcal{X})$ defined in $\{w\}$. $d_{safe}$ is the safety distance. The number of constraints in Eqn. \eqref{eqn:soa} increases the computational complexity significantly. The problem can be simplified further by considering the bounding circles of the projected vertices of the MMR base, manipulators, and the object. This collision geometry reduces the number of constraints and the computational complexity. We have implemented circumscribing bounding circles for each MMR base, manipulator, and object in the ground plane. The center of the circles for the $i$-th MMR base, manipulator, and the object in the ground are located at $p_{base,i}, p_{arm,i}$ and $p_{obj,i}$ with radius $r_{base,i}, r_{arm,i}$ and $r_{obj,i}$. The cyan, purple, and gray area in Fig. \ref{fig:soa} shows the circumscribing circles for the $i-th$ MMR base, manipulator, and object. The reduced static collision avoidance constraints are in the Eqn. \eqref{eqn:soar}

\begin{equation}
	\begin{aligned}
		\mathbf{A}p^k_{m,i}\leq \boldsymbol{b} - d_{safe} - r_{m,i}, \mathrm{P}:(\mathbf{A},\boldsymbol{b})\\
		\forall m \in \{base,obj,arm\}, \forall i \in [1,n], \forall k \in [1,N_h]
	\end{aligned}
	\label{eqn:soar}
\end{equation}
 where $d_{safe}\in \mathbb{R}$ is the safety distance.
\subsubsection{Dynamic Obstacle Avoidance} \label{sec:doa}
The formation $\mathcal{F}$ must not collide with any of the dynamic obstacles $\mathcal{O}_{dyn}$. The space occupied by the formation $\mathcal{B}(\mathcal{X})$ should not overlap with the dynamic obstacles i.e. $\mathcal{B}(\mathcal{X}) \cap \mathcal{O}_{dyn} = \emptyset$. We implement the dynamic obstacle avoidance by introducing a nonlinear constraint between the collision geometries of the formations and the dynamic obstacles. The collision geometry of the dynamic obstacles are considered as circles with radius $r_{dyn,d},\ \forall d\in [1,n_{dyn}]$ located at $p^k_{dyn,d}, \forall k \in [1, N_h]$ in the ground plane, where $d \in [1, n_{dyn}]$ represents the $n_{dyn}$ number dynamic obstacles' ($\mathcal{O}_{dyn}$) index and  represents the number of dynamic obstacle present at the beginning of planning horizon. The position $p^k_{dyn,d} = p_{dyn,d}+v_{dyn,d}kT_c,\ \forall d,\ \forall k \in [1,N_h]$ is predicted with the position $p_{dyn,d}$ and velocity $v_{dyn,d}$ measured at the beginning of each planning horizon using camera, LiDAR based perception system or a motion capture system. Any other dynamic obstacle state estimation model would work with the proposed motion planning algorithm. The accuracy of the estimation model impacts the collision avoidance behavior.
 
The same collision geometry defined for the static obstacle avoidance in Section \ref{sec:soa} for the base, manipulator of the MMRs, and the object are utilized here. The constraints in Eqn. \eqref{eqn:doa} ensure that the obstacle does not intersect with the formation.

\begin{equation}
    \begin{aligned}
        ||p^k_{dyn,d}-p^k_{m}|| \geq r_{dyn,d} + r_m + d_{safe,dyn}\\
        \forall m \in \{base,obj,arm\}, \forall d \in [1,n_{dyn}], \forall k \in [1,N_h]
    \end{aligned}
    \label{eqn:doa}
\end{equation}

\subsubsection{Self Collision Avoidance} \label{sca}
For collision avoidance with the object and the other MMRs, the $i$-th MMR needs to be within the convex wedge shown in Fig. \ref{fig:soa} defined by two infinite vertical planes $\mathcal{H}_i$ and $\mathcal{H}_{(i+1)\%n}$ as shown in Fig. \ref{fig:soa}. The convex wedge specifies the workspace for the $i$-th MMR free from movements of the neighboring MMRs. The planes $\mathcal{H}_i$ and $\mathcal{H}_{(i+1)\%n}$is define considering the workspace of MMRs. Here, we have equally divided the space around the periphery of the object, starting at the CoM of the object for each MMR, as the grasping point is equispaced. The vertical plane $\mathcal{H}_i$ is defined as follows.

\begin{equation}\label{eqn:13}
\mathcal{H}_i =\{\textrm{x}\in\mathbb{R}^2:H_i\textrm{x}\leq h_i, H_i\in\mathbb{R}^{1\times 2}\}, 0\leq z \leq + \infty
\end{equation}

The self-collision avoidance for the $i^{th}$  MMR is defined in the following
\begin{equation}\label{eqn:sca}
		H_i\mathrm{v}_{<x,y>}\leq h_i, H_{(i+1)\%n}\mathrm{v}_{<x,y>}\geq h_{(i+1)\%n}, \forall  \mathrm{v} \in \mathscr{V}_i(q_i)
\end{equation}

where $\mathscr{V}_i(q_i)$ is the set of vertices of the $i$-th  MMR.
\subsubsection{Grasp Constraints} \label{gc}
The MMRs grasp the object at its periphery at equal spacing from each other to have equal workspace among them and for optimal wrench interaction with the object.
There should be no relative movement between the EE and the grasped object throughout the task to ensure stable formation as the object is rigid. The grasp constraint for $i$-th MMR is 
\begin{equation}\label{gpoc}
 p_{ee,i}=p +{}^w_o\textrm{R}(\psi)^or_i, \phi_{ee,i} = \psi
\end{equation}

where $p_{ee,i}$ is $i$-th MMR's EE position, ${}^w_o\textrm{R}$ is the rotation matrix between $\boldsymbol{o}$ and $\boldsymbol{w}$. We represent Eqn. \eqref{gpoc} by $g_i(\mathcal{X}) = 0$.
\section{Object Transportation}\label{sec:Results}
We validate the proposed motion planning algorithm in simulation and hardware experiments with the nonholonomic MMRs that accept velocity as a control input. The system dynamics for the MMRs are approximated using the fourth-order Runge-Kutta method as mentioned in Eqn. \eqref{stf}. The NMPC problem of the local motion planning is solved using the CasADi package \cite{2019_Andersson} with an Interior point optimization (Ipopt) method.

\subsection{Simulation}
The MMRs with a differential drive mobile base have the same forwarding and reversing capabilities. The  Denavit-Hartenberg (DH) parameters are mentioned in Table \ref{tab:0}.

\begin{table}[htbp]
	\caption{DH Parameters Value for the manipulators}
	\label{tab:0}
	\begin{center}
		\begin{tabular}{|c|c|c|c|c|}
			\hline
			\textbf{Joint} & $d\ (m)$ & $a\ (m)$ & $\alpha\ (rad)$& $\theta\ (rad)$\\
			\hline
			Joint 1 & 0.070 & 0 & 0 & $q_{a,1}$\\
			\hline
			Joint 2 & 0 & 0 & $0.5\ \pi$ & $q_{a,2}$\\
			\hline
			Joint 3 & 0.100 & 0 & $- \pi$ & $q_{a,3}$\\
			\hline
			Joint 4 & 0.125 & 0 & $ \pi$ & $q_{a,4}$\\
			\hline
			Joint 5 & 0 & 0.120 & $-0.5\ \pi$ & $q_{a,5}$\\
			\hline
			Gripper & 0 & 0 & 0 & 0\\
			\hline
		\end{tabular}
	\end{center}
\end{table}
We select the operational velocity of the formation $v_{op} = 0.15\ m/s$ and use prediction horizon time $T_h = 9\ s$, trajectory execution time $T_e = 3\ s$ and the discretization time step $T_c = 0.25\ s$. The safety margins  $d_{safe} = 0.05\ m$ and $d_{safe,dyn} = 0.1\ m$ for static and dynamic obstacle avoidance to keep the formation safe during object transportation. A higher margin restricts the formation from nearing the obstacles and hence reduces the obstacle-free space.
The tuned optimization weights are $\mathbf{W_u} = diag(repeat([0.05,0.25,2.5,2.5,2.5,5,2.5],5))$, $\mathbf{W_e} = diag([0.01,0.01])$ and $\mathbf{W_{N_h}} = 10^5$. We assign a lower weight to the base, prioritize the use of the mobile base motion over the arm, and avoid joint limits when reaching for the arm. The base angular motion is assigned a relatively higher weight than linear to reduce rotation and hence improve ground trajectory smoothness. One of the manipulator joints is assigned a higher weight to reduce the more frequent motion in the $\pm z$ direction. The lower weight to the global reference tracking error ensures flexibility for a smooth trajectory with better dynamic obstacle avoidance behavior.

\begin{figure}[h]
	
	\begin{subfigure}[t]{0.24\textwidth}
		\includegraphics[width=115px]{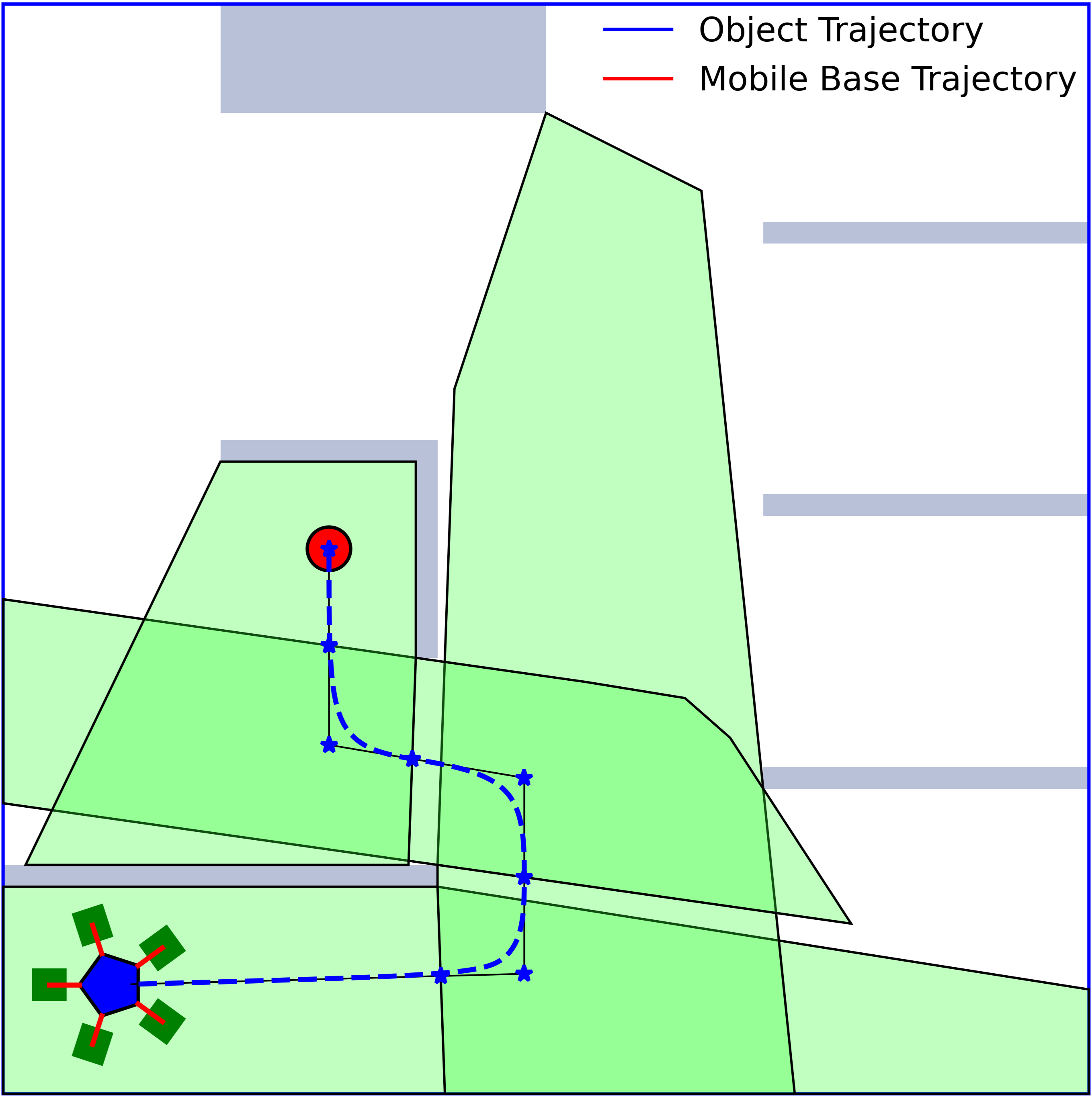}
		\subcaption{\label{fig:sima} $t = 0\ s$}
	\end{subfigure}
	\begin{subfigure}[t]{0.24\textwidth}
		\includegraphics[width=115px]{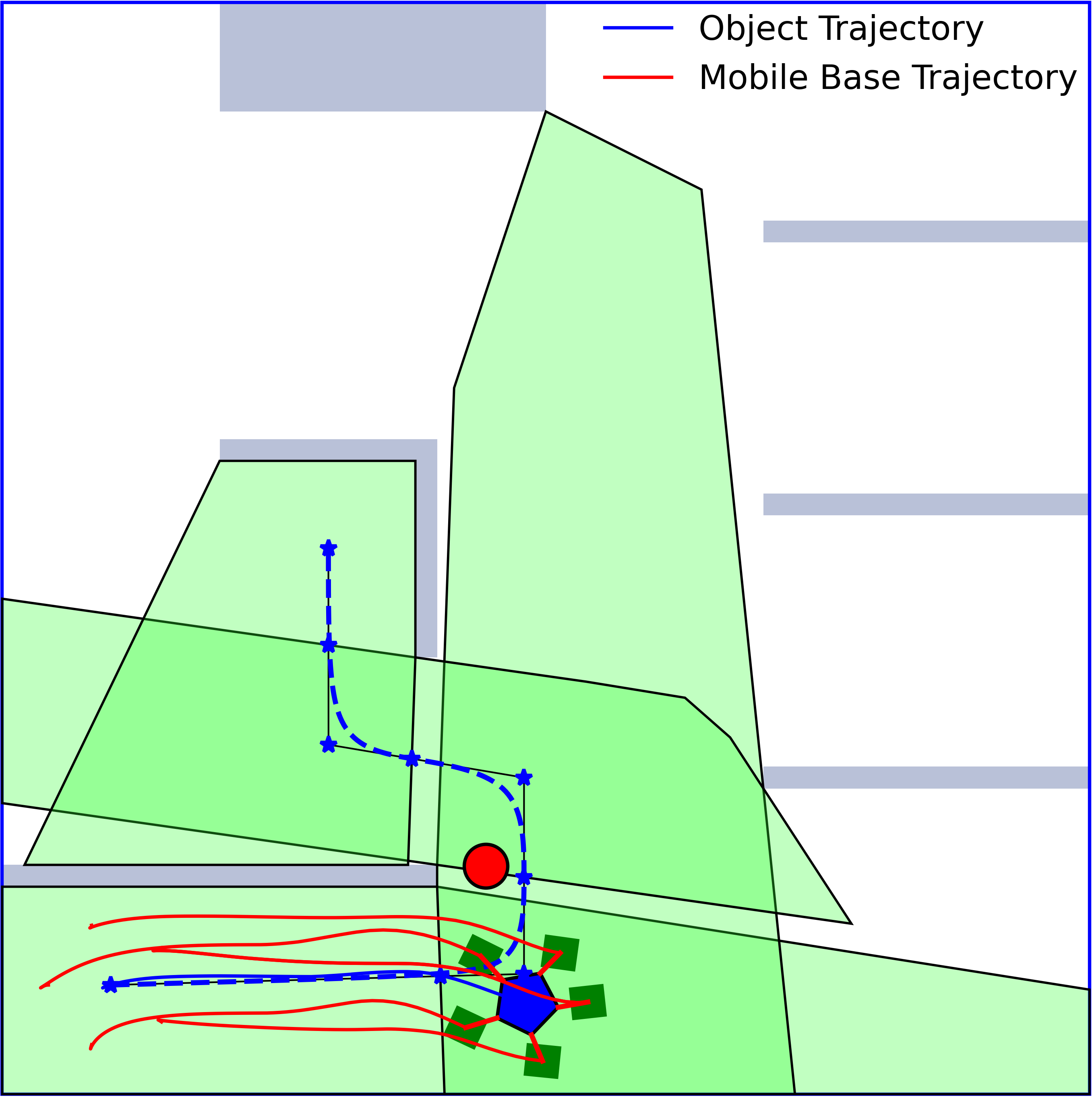}
		\subcaption{\label{fig:simb} $t = 32.5\ s$}
	\end{subfigure}
	\begin{subfigure}[t]{0.24\textwidth}
		\includegraphics[width=115px]{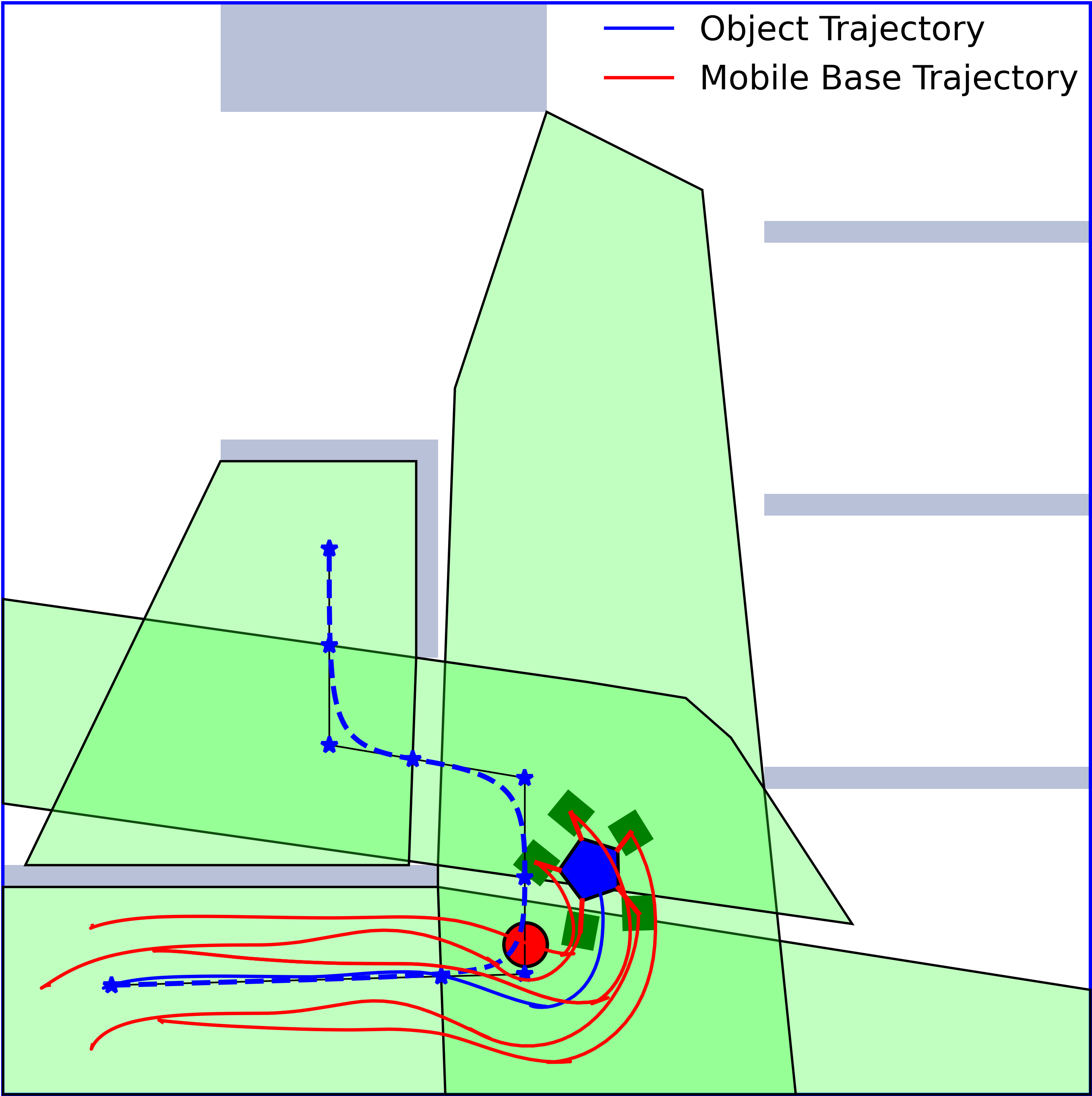}
		\subcaption{\label{fig:simc} $t = 40.5\ s$}
	\end{subfigure}
	\begin{subfigure}[t]{0.24\textwidth}
		\includegraphics[width=115px]{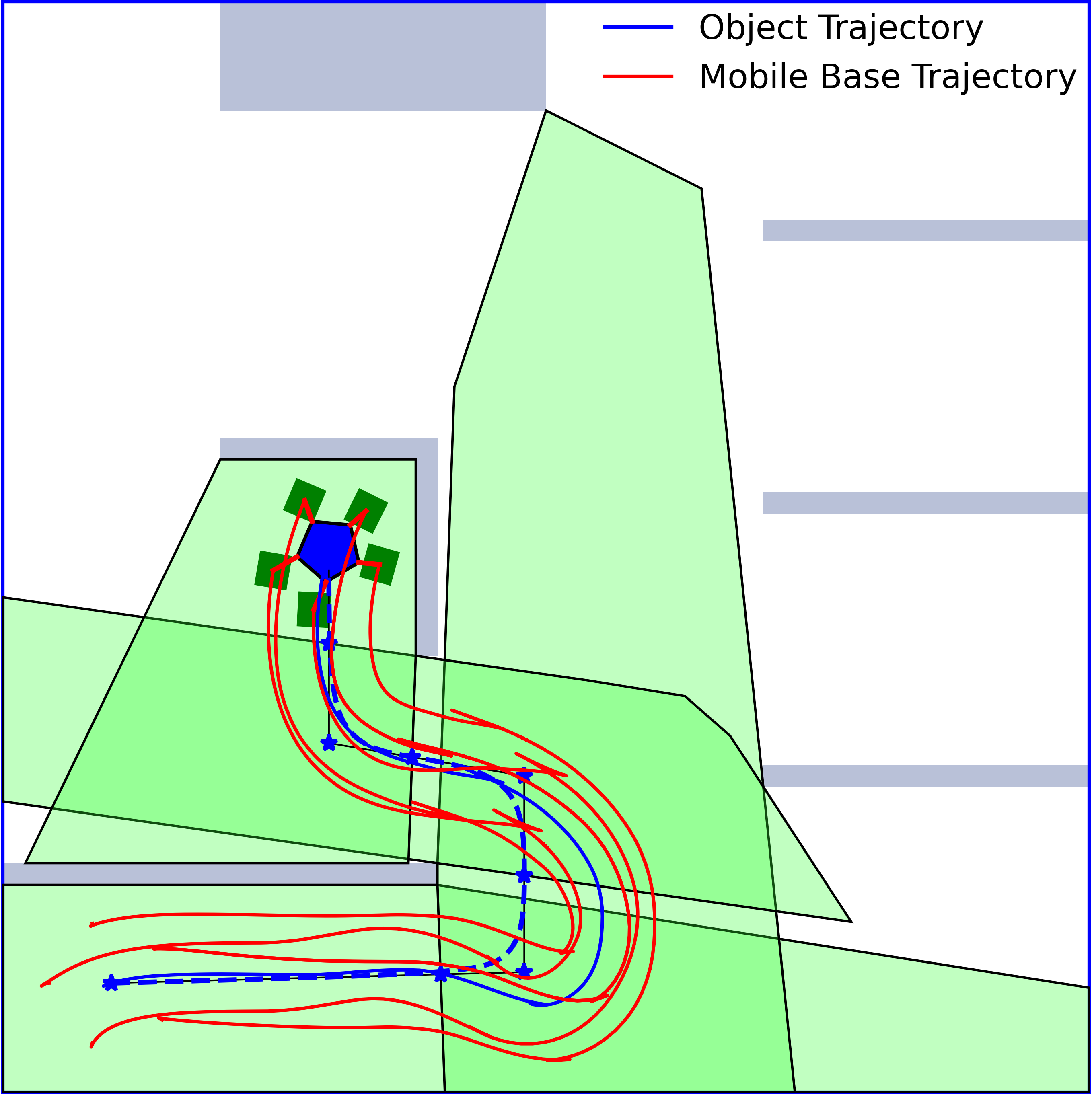}
		\subcaption{\label{fig:simd} $t = 79\ s$}
	\end{subfigure}
	\begin{subfigure}[t]{0.48\textwidth}
		\includegraphics[width = 235px]{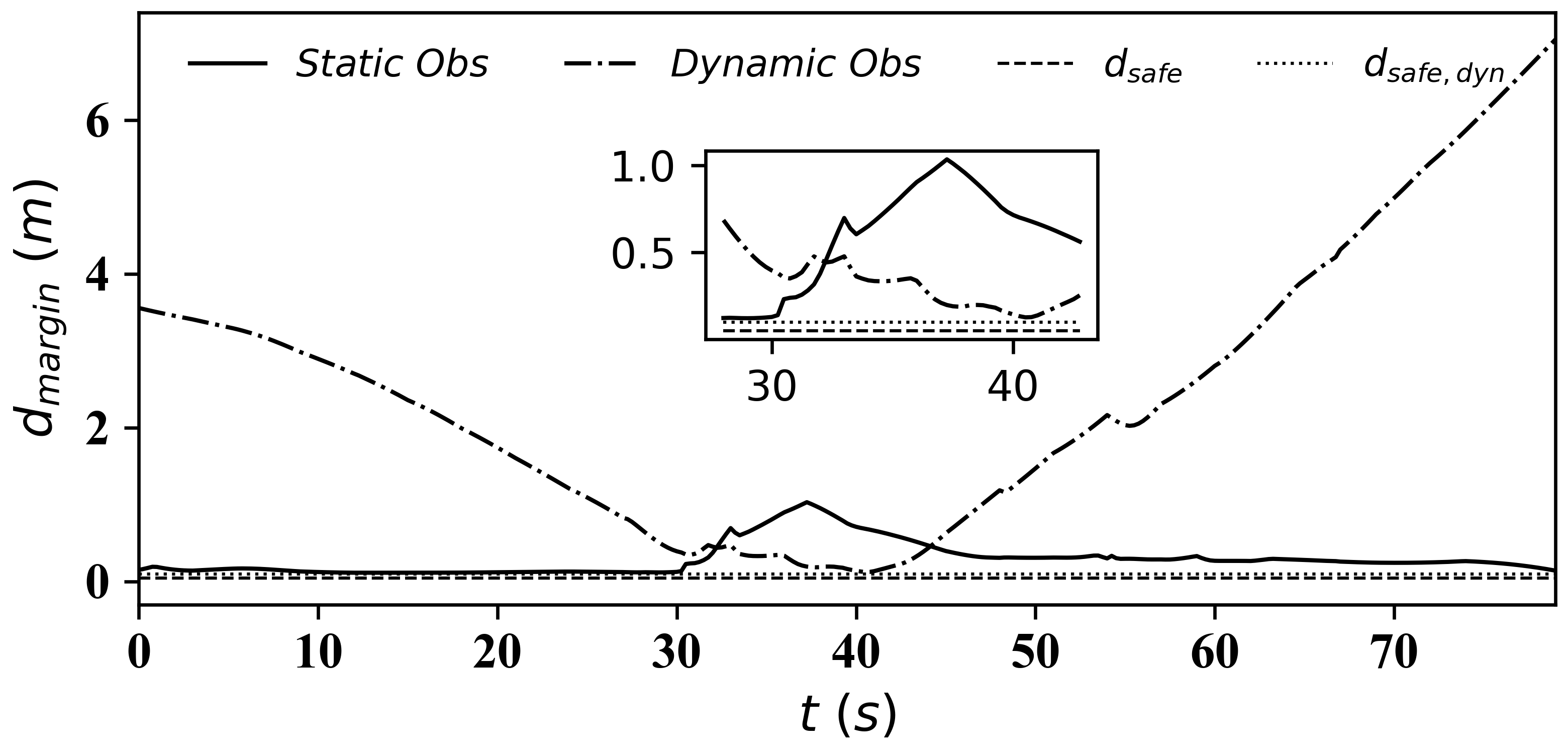}
		\subcaption{\label{fig:SafetyMargin} Safety margins. }
	\end{subfigure}
	\caption{The snapshots of object transportation from a start (Fig. \ref{fig:sima}) to a goal (Fig. \ref{fig:simd}) by five MMRs in $10m\times10m$ environment. The red circle indicates dynamic obstacle is in its current state. The horizontal lines in Fig. \ref{fig:SafetyMargin} plots static and dynamic safety threshold $d_{safe}\ =\ 0.05\ m$ and $d_{safe,dyn}\ =\ 0.1\ m$.}
	\label{fig:Sim}
\end{figure}

The five MMRs (the mobile base in deep green rectangle and manipulator's arm in red line) start transporting (Fig. \ref{fig:sima}) the object while grasping at its periphery through a narrow corridor of $1.9m$ in an environment of size $10m\times10m$ . While the MMRs come out of the corridor, it encounters dynamic obstacles in Fig. \ref{fig:simb}, while taking a sharp left turn. The generated motion plan successfully navigates the formation, avoiding dynamic obstacles, and turns toward (Fig. \ref{fig:simc}) the goal. The MMRs successfully transport the object through the narrow doors and complete the task without any collision (Fig. \ref{fig:simd}). Fig. \ref{fig:SafetyMargin} plots the shortest distance $d_{margin}$ from the formation to any static and dynamic obstacles. The $d_{margin}$ in Fig \ref{fig:SafetyMargin} for the static and the dynamic obstacles being always positive indicates successful collision avoidance behavior.

\subsection{Hardware Experiments}
\begin{figure}[htbp]
	\centerline{\includegraphics[width = 210px]{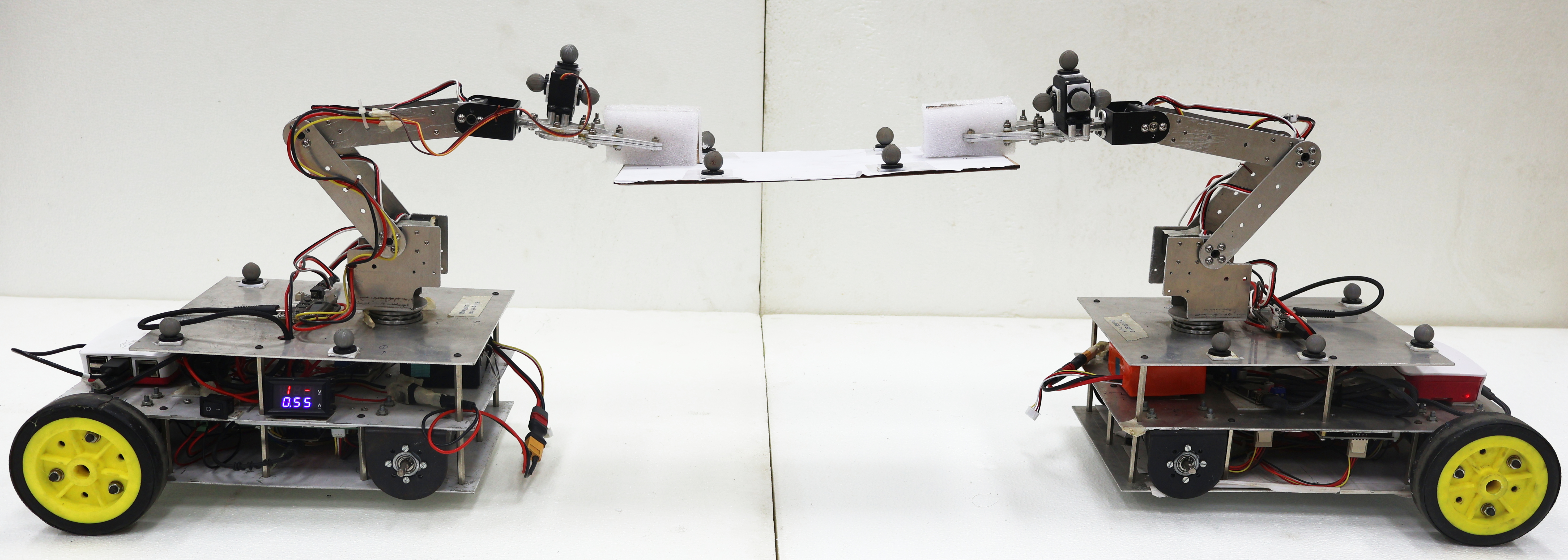}}
	\caption{Experimental Setup of two in-house developed nonholonomic MMRs.}
	\label{fig:NonHoloSetup}
\end{figure}

\begin{figure}[htbp]
	\begin{subfigure}[t]{0.24\textwidth}
		\includegraphics[width=115px]{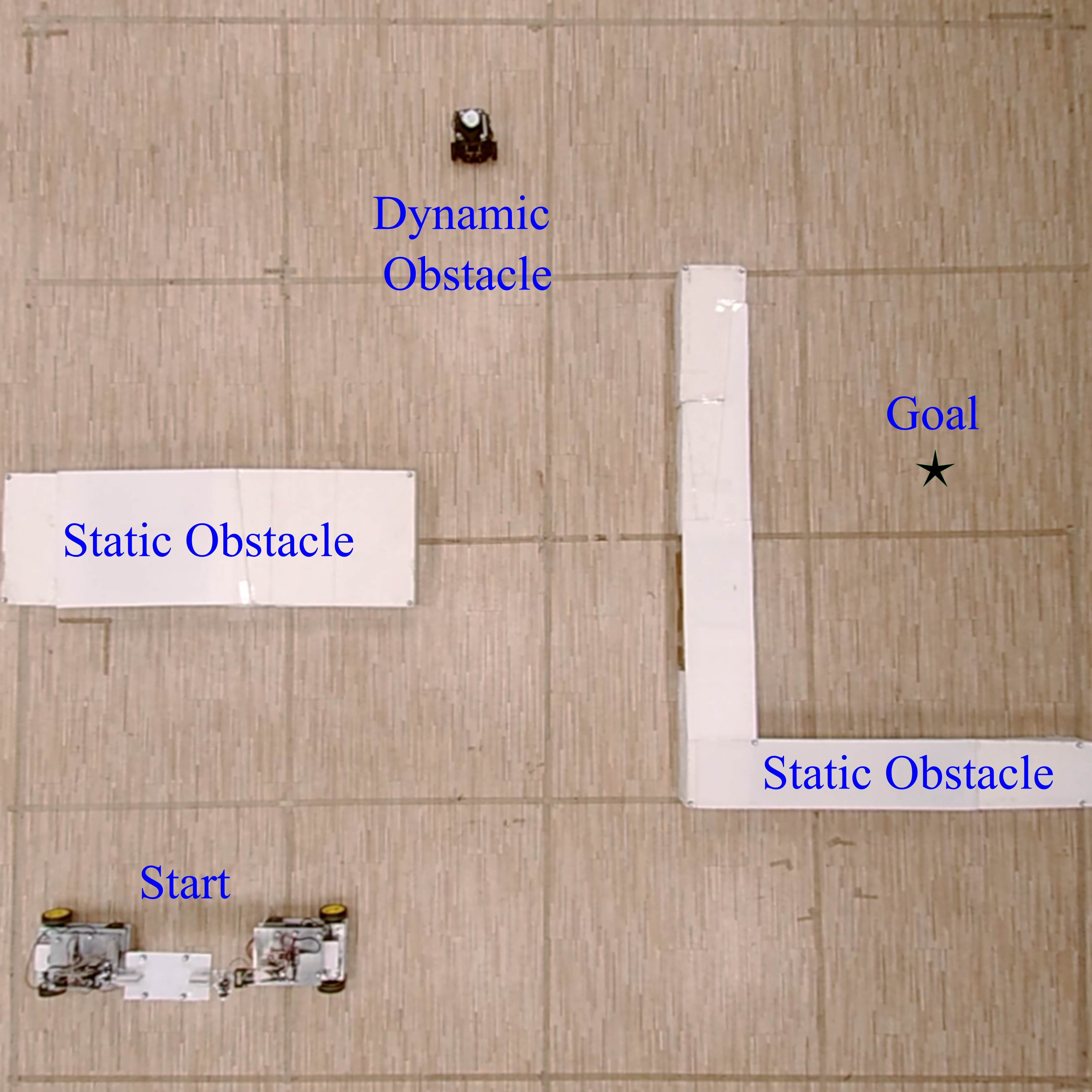}
		\subcaption{\label{fig:expa} $t = 0\ s$}
    \end{subfigure}
	\begin{subfigure}[t]{0.24\textwidth}
		\includegraphics[width=115px]{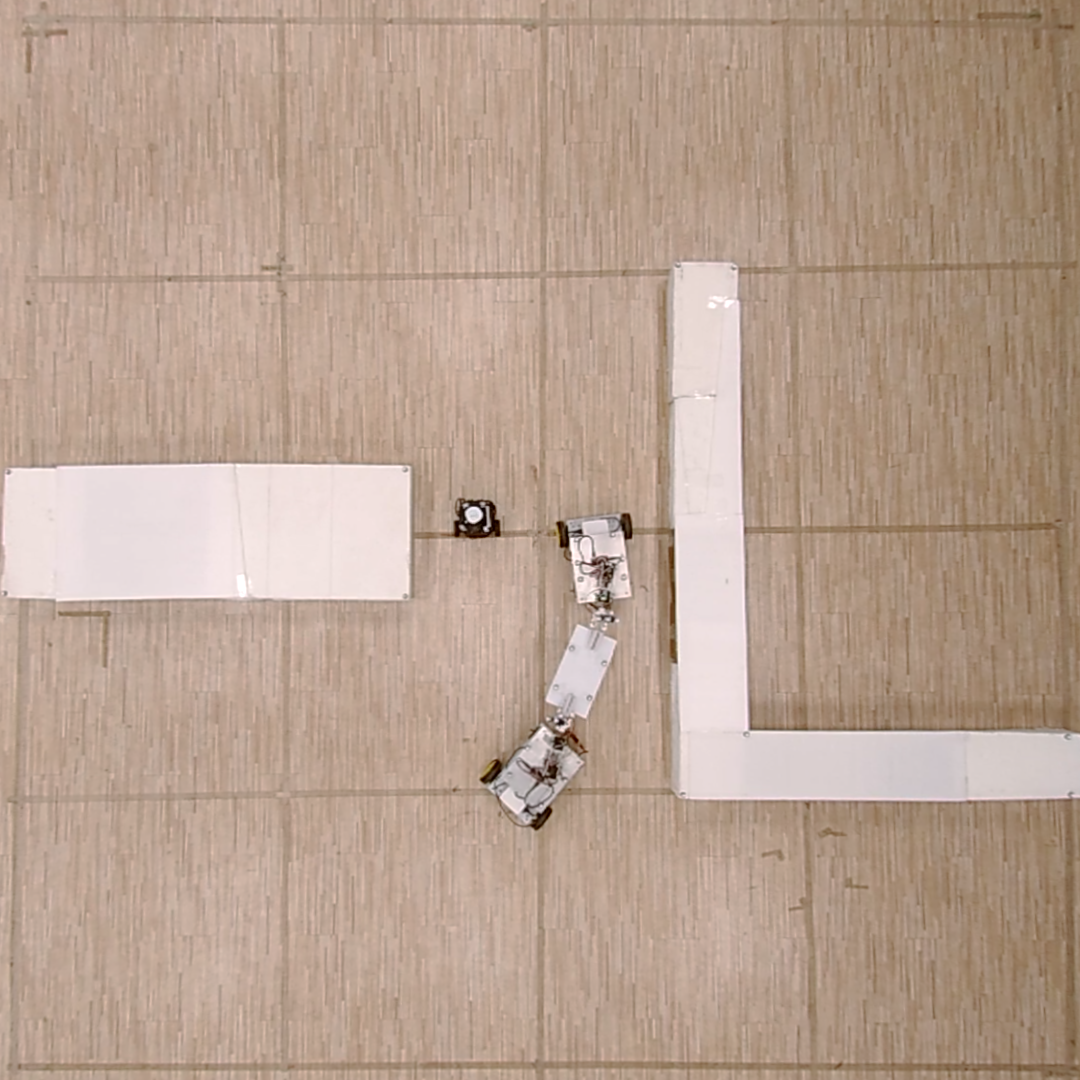}
		\subcaption{\label{fig:expb} $t = 15.10\ s$}
    \end{subfigure}
	\begin{subfigure}[t]{0.24\textwidth}
		\includegraphics[width=115px]{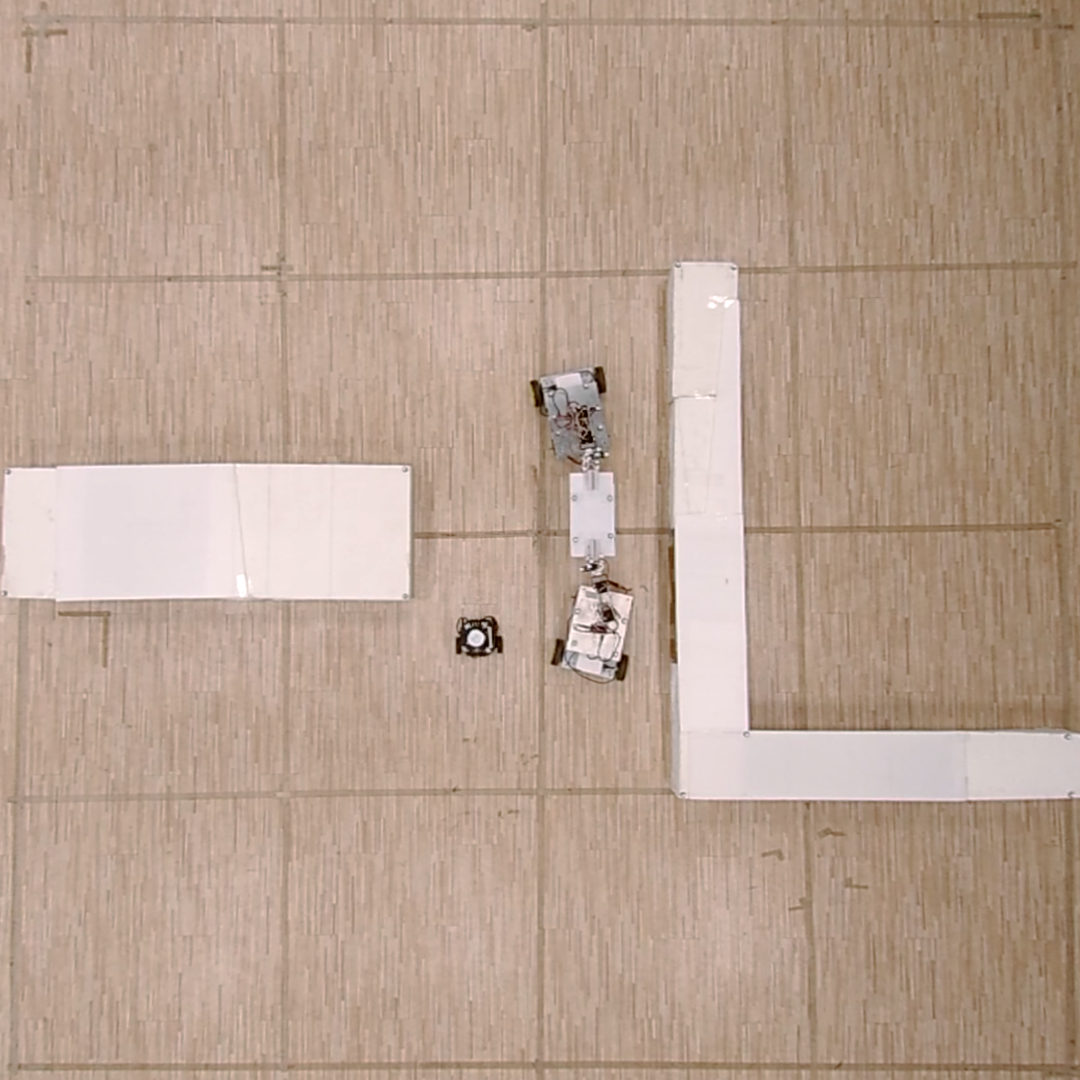}
		\subcaption{\label{fig:expc} $t = 20\ s$}
    \end{subfigure}
	\begin{subfigure}[t]{0.24\textwidth}
		\includegraphics[width=115px]{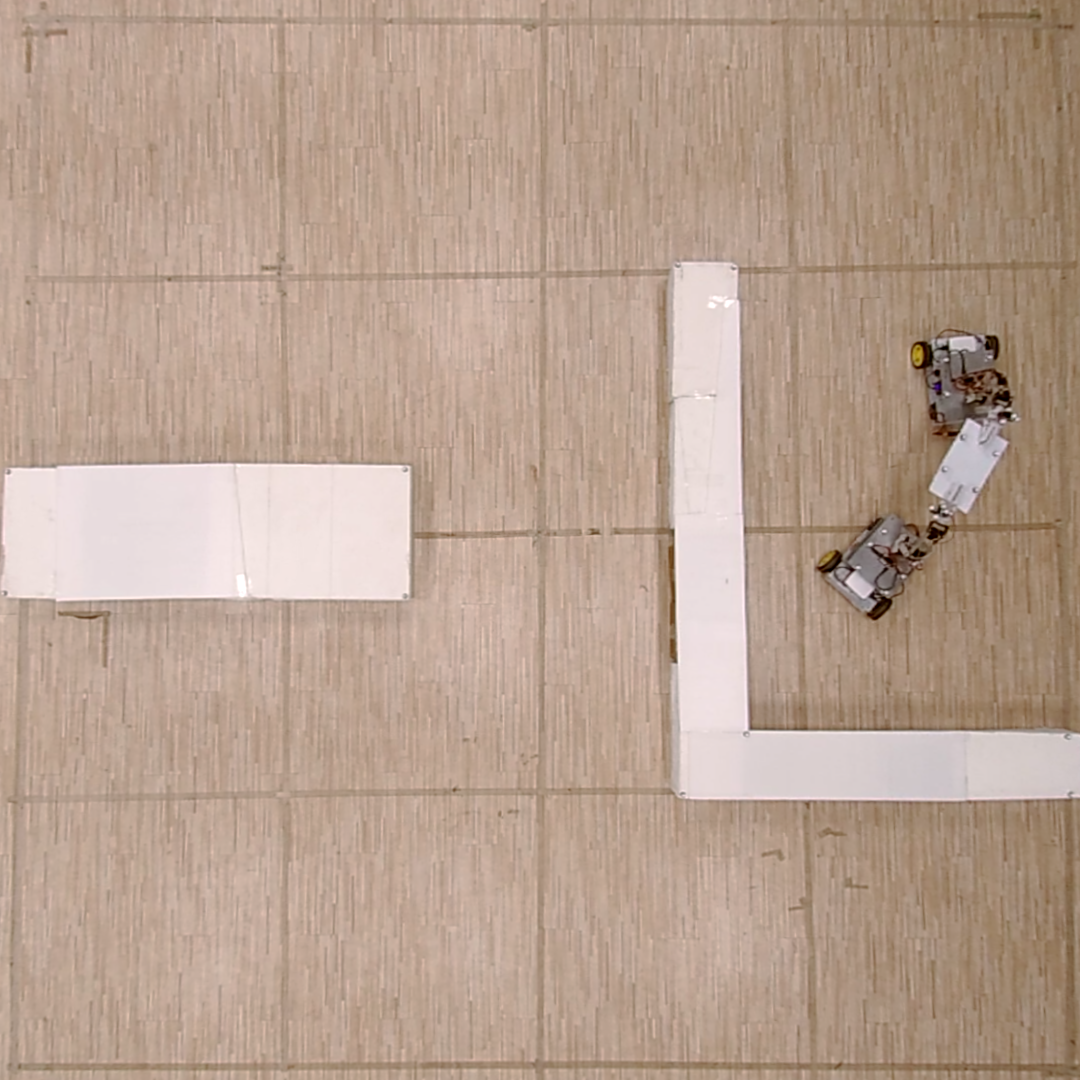}
		\subcaption{\label{fig:expd} $t = 47\ s$}
	\end{subfigure}
	\caption{Two MMRs transport the rectangular object. The MMRs encounter a dynamic obstacle and started avoidance maneuver (Fig. \ref{fig:expb}). It successfully avoids the dynamic obstacle \ref{fig:expc}) and reaches the goal point \ref{fig:expd})}
	\label{fig:ExpSnap}
\end{figure}

We perform experiments with our in-house developed ROS-enabled MMRs to evaluate the motion planning algorithm in Section \ref{OMP} in an environment ($4\ m\times4\ m$) with static and dynamic obstacles. We have defined a specific path for the dynamic obstacles (turtle bot). However, the path of the obstacle is not known to the mobile manipulators. The nonholonomic MMR bases are made of two disc wheels each separately driven by geared motor with an encoder. Fig. \ref{fig:NonHoloSetup} shows two MMRs both grasped an object to transport it in an indoor environment shown in Fig. \ref{fig:expa}.

The manipulator of the MMRs shown in Fig. \ref{fig:NonHoloSetup} is same as the manipulator used for the simulation, described in Table \ref{tab:0}, except for the joint 5. We have removed and fixed the joint 5 because of the joint 2 torque limitations. The adjusted DH parameters of the gripper are $d = 0.120\ m, a = 0, \alpha = 0$, and $\theta = 0$.
The planned trajectory and the control input for the MMRs by the online motion planner (Section \ref{OMP}) are sent to the respective MMRs. The trajectory tracking controllers for the mobile base and manipulator uses online motion planner's computed control input as feed forward and PID feedback controller to ensure desired trajectory tracking. It uses external motion capture system data and joints encoder data for the mobile base and manipulator feedback.

Fig. \ref{fig:ExpSnap} shows the snap of the object transport from the start (Fig. \ref{fig:expa}) to the goal (Fig. \ref{fig:expd}). It encounters a dynamic obstacle and start avoidance maneuver. Fig. \ref{fig:expb} shows when the formation approaches the dynamic obstacle and finally avoids (Fig. \ref{fig:expc}) the obstacle to reach the goal (Fig. \ref{fig:expd}).

\begin{figure}[htbp]
	\centerline{\includegraphics[width = 240px]{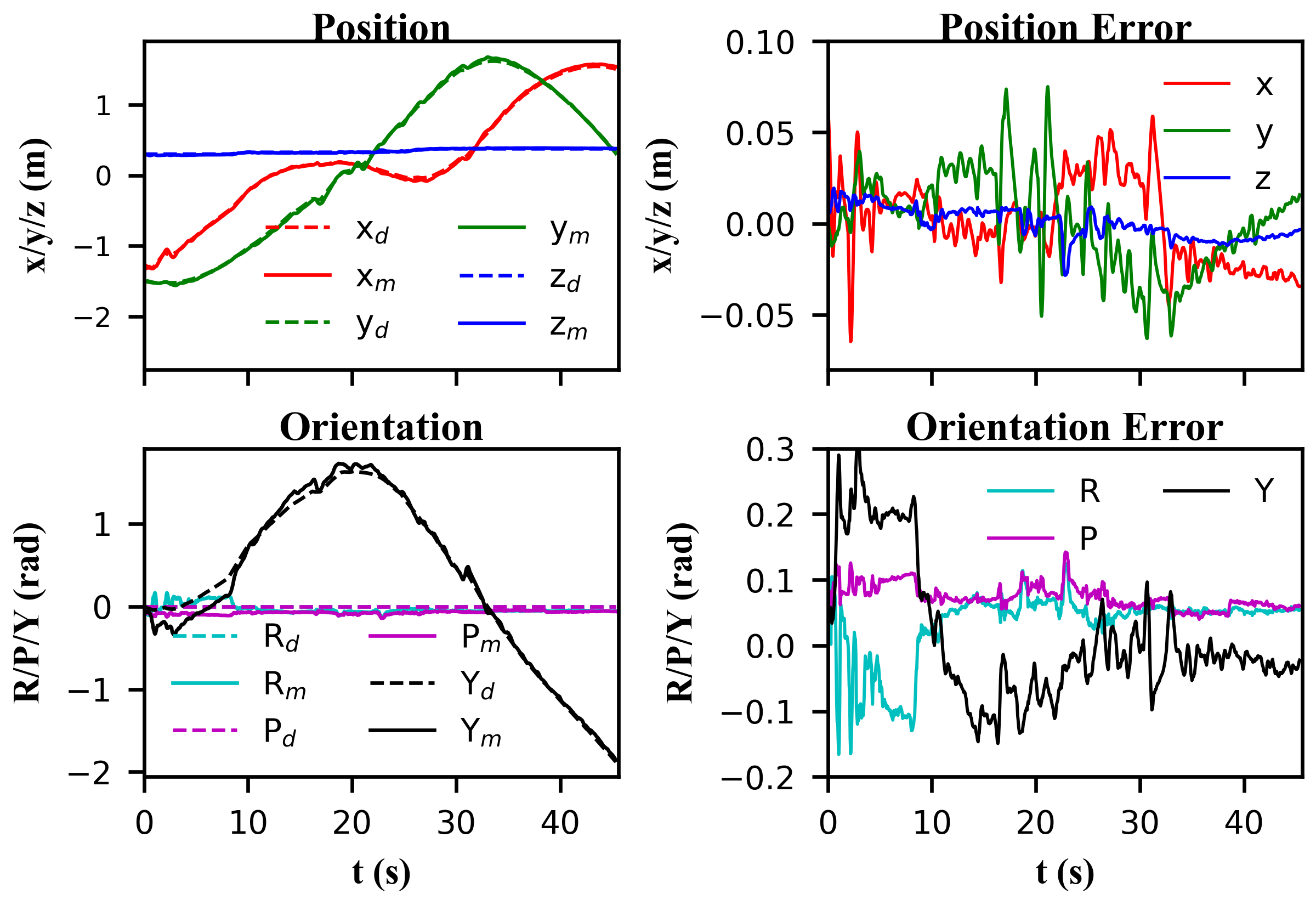}}
	\caption{Trajectory of the CoM of the object. The subscript d and m of the legend represents the planned and actual values.}
	\label{fig:obj_pose}
\end{figure}

\begin{figure}[htbp]
	\centerline{\includegraphics[width = 210px]{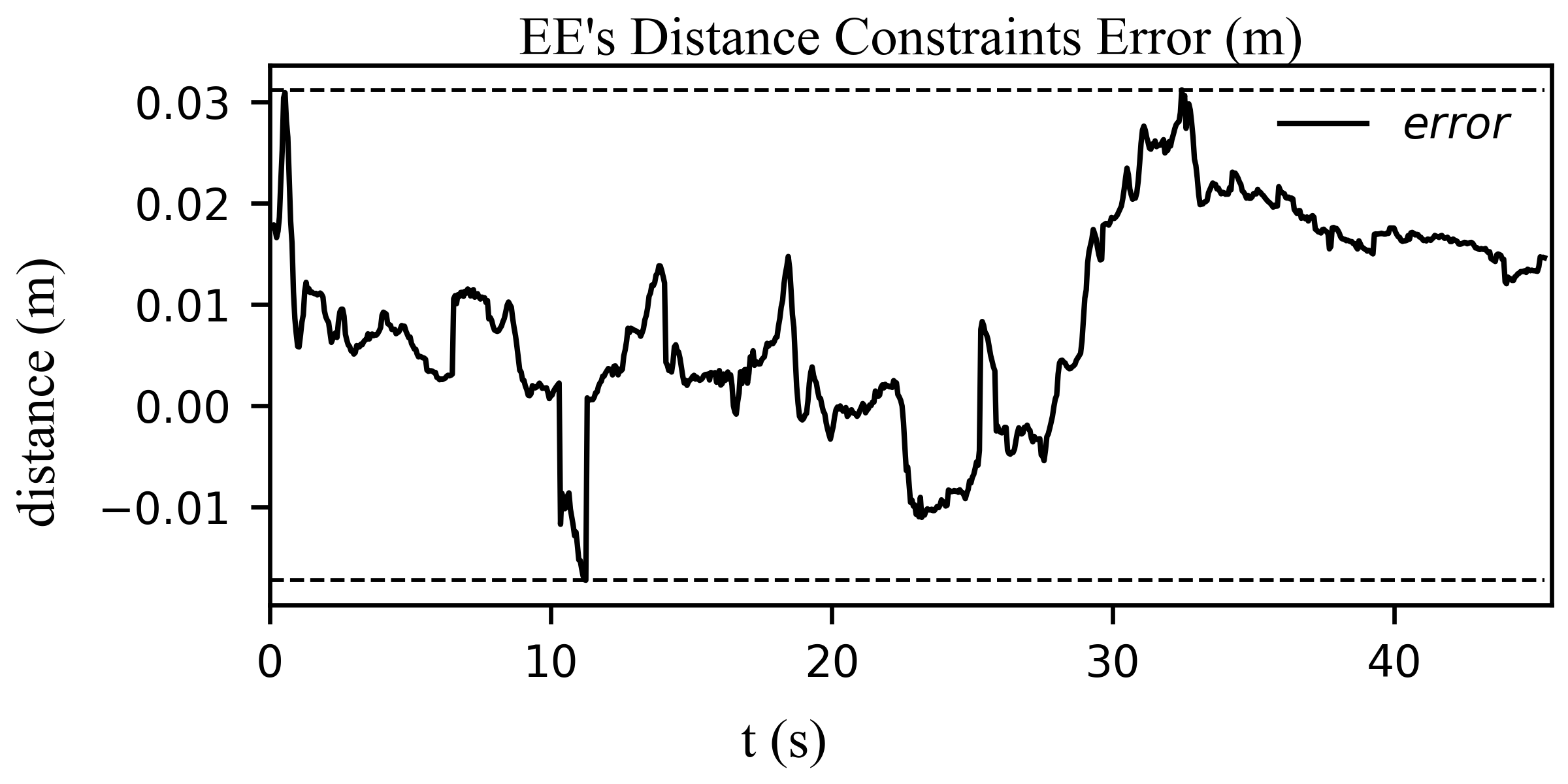}}
	\caption{The distance between the two EE during the object transportation.}
	\label{fig:ee_constr}
\end{figure}

Fig. \ref{fig:obj_pose} shows the planned and the actual trajectory of the CoM of the transported object. The position error remains within $0.05\ m$, and the orientation error remains within $0.15\ rad$.
The  $z$  height increases near $t = 10\ s$ and $t = 28\ s$ before taking sharp turn to reduce the inter robot distance and turning radius. The error in fixed distance between the EEs' grasping point in Fig. \ref{fig:ee_constr} shows that the coordination between the MMRs has been maintained.

\subsection{Comparison} \label{ComparativeAnalysis}
We present a comparison of performances of our proposed path planning algorithm for the environment shown in Fig. \ref{fig:5e} and motion planning algorithms for a simplified environment where all the three algorithm (proposed, \cite{2017_AlonsoMora, 2024b_Keshab}) works. We have run all the algorithms in Python on a Laptop equipped with an AMD Ryzen 5800H CPU and 16 GB RAM. The comparison results are presented as follows.
 
\textit{Path Planner:} We compare our proposed path planning algorithm with the IRIS-based algorithms \cite{2017_AlonsoMora, 2024b_Keshab} and a RRT Connect like technique \cite{2022_Zhang} (computes only path) for the environment shown in Fig. \ref{fig:5e}.  Table \ref{tab:PathLengthTime} shows the comparison of path length and the computation time for ten runs of each algorithm.

\begin{table}[h]
	\centering
	\caption{Path Planner Comparison.}
	\begin{tabular}{@{}lcc@{}}
		\hline
		&Path Length & Computation Time\\
		& $\text{mean} \pm \text{sd}\ (m)$
		& $\text{mean} \pm \text{sd}\ (s)$ \\
		\hline
		Proposed&$9.22 \pm 0$&$0.09 \pm 0.03$\\
		\textit{Keshab et al.}\cite{2024b_Keshab} &$9.87 \pm 0$&$3.74 \pm 1.60$ \\
		\textit{Alonso-Mora et al.}\cite{2017_AlonsoMora}&$12.09 \pm 3.88$&$8.36 \pm 6.28$\\
		\textit{Zhang et al.}\cite{2022_Zhang} & $12.61 \pm 4.20$ & $35.01 \pm 14.02$ \\
		\hline
	\end{tabular}
	\label{tab:PathLengthTime}
\end{table}
The convex optimization-based path planning approaches in \cite{2014_Deits, 2017_AlonsoMora} rely on randomly generated seed points to compute the path and its corresponding obstacle-free convex region. Such random seeding coupled with convex optimization incurs high computational cost. The targeted seeding strategy proposed in \cite{2024b_Keshab} shows improved performance relative to fully sampling-based techniques \cite{2017_AlonsoMora,2022_Zhang}. Our method deterministically generates the shortest feasible path and analytically computes the associated obstacle-free convex region thereby reduces the computational time.

\textit{Motion Planner:} We compare the computational time and control effort of our proposed online motion planning technique with the holonomic MMRs' planning algorithm proposed by \textit{Keshab et al.} \cite{2024b_Keshab} and  \textit{Alonso-Mora et al.} \cite{2017_AlonsoMora}. All algorithms plan the motion of two MMRs in a dynamic environment with a dynamic obstacle with environment details mentioned in Section 3.1 of the \href{https://drive.google.com/file/d/1lmj1ZvdymSjk0P-nYwg_vj7pajnEIXTU/view?usp=sharing}{attached file}. We compare our nonholonomic planning framework against holonomic baseline, as to the best of our knowledge there exist no other nonholonomic multi-robot collaborative motion planners that address dynamic obstacle avoidance. The MMRs are the same except for the base motion constraints: nonholonomic and holonomic. Computational time is measured for each local planning horizon. The control effort is computed for a complete trajectory. As shown in the Table \ref{tab:ComputationTime}, the computation time and the control efforts of our method are lower than the algorithm in \cite{2017_AlonsoMora} and slightly higher than the algorithm proposed in \cite{2024b_Keshab}. The nonholonomic constraint for the mobile base reduces the solution space compared to its holonomic counterpart, resulting in increased computation time and control efforts. The proposed method achieves real-time operation in Python and we anticipate substantially faster runtime with C++ implementation.

\begin{table}[h]
	\centering
	\caption{Motion Planner Comparison.}
	\begin{tabular}{@{}lccc@{}}
		\hline
		&\multicolumn{3}{c}{Computation Time (s)}\\
		\hline
		  & min& $\text{mean} \pm \text{sd}$ & max\\
		\hline
		Proposed&$0.199$&$0.580 \pm 0.231$ & $1.855$\\
        \textit{Keshab et al.}\cite{2024b_Keshab} &$0.227$&$0.272 \pm 0.035$ & $0.346$\\
		\textit{Alonso-Mora et al.}\cite{2017_AlonsoMora}&$0.46$ & $0.857 \pm 0.256$ & $1.26$\\
        \hline
        \hline
		&\multicolumn{3}{c}{Control Effort}\\
        \hline
        & $\sum\lvert\lvert\boldsymbol{u}\rvert\rvert_2$ &\multicolumn{2}{c}{$\lvert\lvert\boldsymbol{u}\rvert\rvert_2\ \text{mean} \pm \text{sd}$}\\
        \hline
        Proposed& $64.869$ &\multicolumn{2}{c}{$0.676 \pm 0.468$}\\
        \textit{Keshab et al.}\cite{2024b_Keshab}&$36.136$ &\multicolumn{2}{c}{$0.435 \pm 0.183$}\\
        \textit{Alonso-Mora et al.}\cite{2017_AlonsoMora}&$78.754$ &\multicolumn{2}{c}{$0.984 \pm 0.759$}\\
        \hline
        
	\end{tabular}
	\label{tab:ComputationTime}
\end{table}
\section{Conclusion}\label{sec:Conclusion}
The optimization-free ellipse based polygon computation algorithm is a trade-off, where the polygon suboptimality significantly reduces the computational time. This trade-off does not affect the subsequent NMPC-based planning stage, as the ellipses are aligned with the piece-wise reference path segments and adaptively inflated to ensure adequate obstacle free space for kinodynamic constrained motion and dynamic obstacle avoidance. The proposed path planning technique computes both the path and its associated convex region within $120\ ms$ indicating that the path can be recomputed online during the task execution.

The simulation and experimental results demonstrate that the motion planner generates kinodynamically feasible, collision-free trajectories in dynamic environments in real time, indicating the strong potential for deployment in factory and warehouse like settings. Motion planning for cooperative MMRs during object transportation remains particularly challenging, especially for nonholonomic MMRs due to the kinodynamic constraints and the rigid object-manipulator coupling that must be respected during task executions.

Our simulations and hardware experiments indicate that the trajectory may exhibits non-smooth transitions at the intersection of the obstacle-free convex polygon. However, this did not adversely affect the experiments as planner ensures the controls constraints and the low-level tracking controller effectively managed traction loss in the mobile base. In the future, we will address the trajectory smoothness at the transitions.


\normalsize
\bibliography{ref}

\section*{Appendix I: Additional Details of Algorithm}\label{sec:ada}
We present some additional implementation details of our proposed Algorithms in Section \ref{sec:ada}, List of Symbols in Section \ref{sec:los}. The analytical ellipse computation technique is described in detail in Section \ref{sec:adaec}, and the dilation of the ellipse is described in Section \ref{sec:adaed}. We add a tabulated list for symbols used in the manuscript in Section \ref{sec:los}.
	
Algorithm \ref{algo:convexify} illustrated convexify simple concave polygon. Here we explain the FindEllipse function in details in Algorithm \ref{algo:FindEllipse} and DilateEllipse in Algorithm \ref{algo:DilateEllipse}.

\subsection*{Ellipse Computation}\label{sec:adaec}
	\begin{figure}[h]
		\centerline{\includegraphics[width =160
			px]{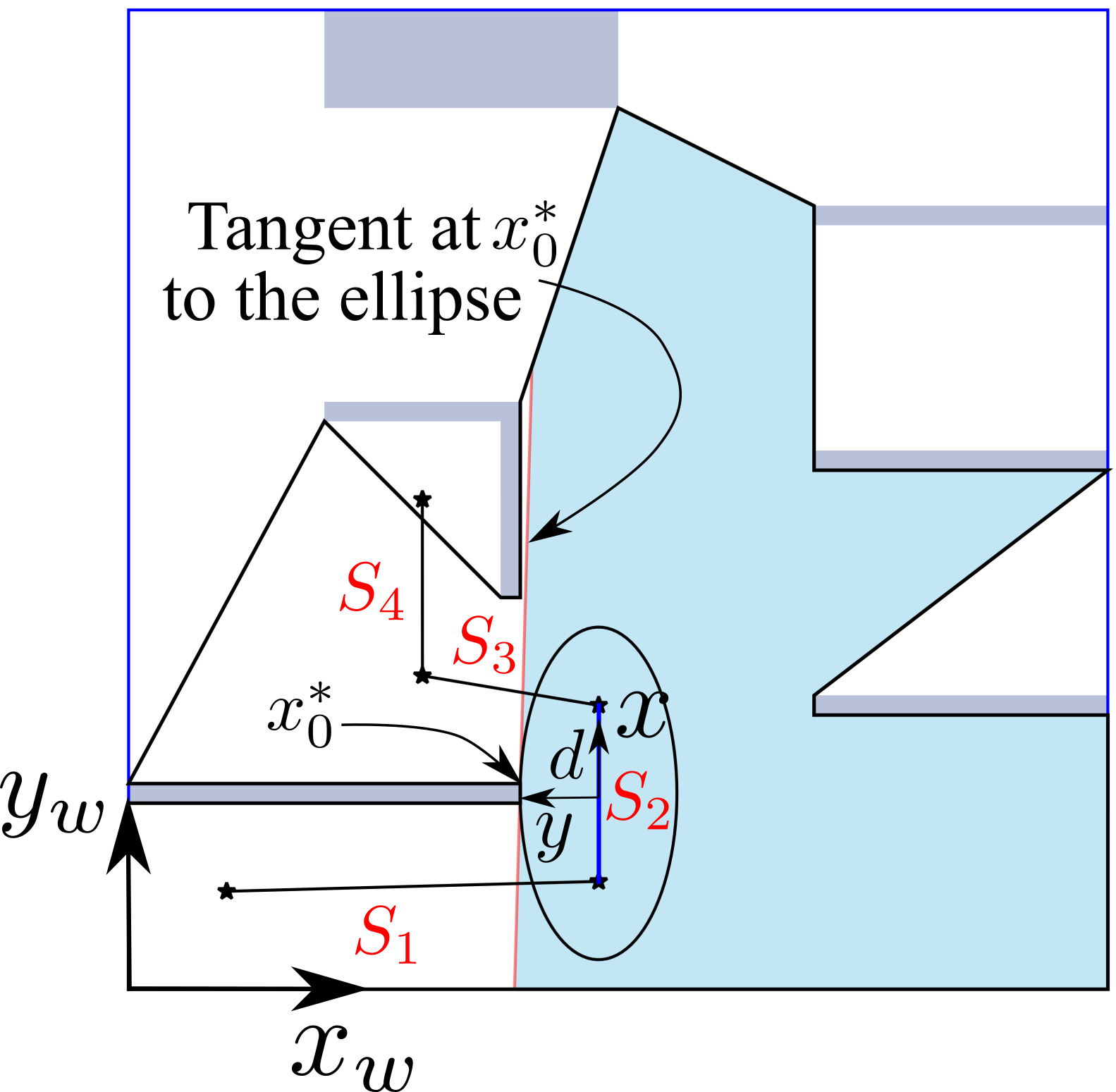}}
		\caption{Ellipse fitting at the center $d$ of the path segment $S_2$.}
		\label{fig:Convexification}
	\end{figure}
	This section describe the ellipse computation method with mathematical details, used in the Polygon Convexification Algorithm \ref{algo:convexify}. We find the Rotation matrix $R$ putting a local frame origin at the center $(d)$ of the ellipse and $x$-axis along the path segment $S_2$ as shown in Fig. \ref{fig:Convexification}. The ellipse computation method has been described in Algorithm \ref{algo:FindEllipse}.
	
	\begin{algorithm}
		\caption{FindEllipse($R,a,\mathrm{d},x^*$)}\label{algo:FindEllipse}
		\begin{algorithmic}[1]
			\Statex \textbf{Input:} Rotation Matrix $R$, Semi-major axis $a$
			\Statex Center of Ellipse $\mathrm{d}$, The nearest concave point $x^*$
			\Statex \textbf{Output:} Ellipse $\kappa(\mathrm{C},d)$
			\Statex\hrulefill
			\State $dist\_x^*$ = Euclidean Distance of $x^*$ from $d$
			\If{$dist\_x^* > (a-\epsilon)$}
			\Comment{$\epsilon\ = 0.001$, when $x^*$ is distant eliminates degeneracy}
			\State $a = max(1.5*dist\_x^*, 2*a)$
			\Comment{Modify $a$ with parameter $1.5, 2$ so $x^*$ can used for ellipse fitting}
			\EndIf
			\State $x^*(x,y) = R(x^* - d)$
			\Comment Converting to local ellipse frame
			\State $b = \lvert x^*_y / \sqrt{1-(\frac{x^*_x}{a})^2} \rvert$
			\Comment Find the minor axis of the ellipse touching $x^*(x,y)$ using $(\frac{x^*_x}{a})^2 + (\frac{x^*_y}{b})^2 = 1$
			\State $\Lambda = diag(a,b)$
			\State $C = R^T \Lambda R$
			\State \textbf{return} $\mathrm{C},d$
			
		\end{algorithmic}
	\end{algorithm}
	
	Once we have $x^*$, the nearest concave vertex to $d$, the distance of $x^*$ from the origin $d$ of the ellipse should be less than the semi-major axis $a$, otherwise one cannot fit the ellipse with $R,a,\mathrm{d},x^*_j$. We check the degeneracy in lines 1-2 and modify $a$ if required. Then we compute the semi-minor axis $b$ using the fundamental ellipse equation for Cartesian coordinates.
	
\subsection*{Ellipse Dilation}\label{sec:adaed} 
	We describe the ellipse dilation method in Algorithm \ref{algo:DilateEllipse}. We dilate the base ellipse computed in the first step of the polygon convexification technique to eliminate the remaining concave vertices. The dilation step length is computed analytically based on the nearest concave vertices of the base ellipse and is described in the Algorithm \ref{algo:DilateEllipse} step-wise.
	\begin{algorithm}[H]
		\caption{DilateEllipse($\kappa_0(C_0, d_0), R, x^*_j$)}\label{algo:DilateEllipse}
		\begin{algorithmic}[1]
			\Statex \textbf{Input:} Base Ellipse $\kappa_0$, The nearest concave point $x^*_j$
			\Statex \textbf{Output:} Ellipse $\kappa_j$
			\Statex\hrulefill
			\State $a, b\gets \kappa_0$
			\Comment Extract semi-major and semi-minor axis from $\kappa_0$
			\State $x^*(x,y) = R(x^*_j - d)$
			\Comment Converting to local ellipse frame
			\State $angle = \tan^{-1}(\frac{x^*_y \cdot a}{x^*_x \cdot b})$
			\Comment{Finding the ellipse angle of vector from $d$ to $x^*_j$ with the semi major axis}
			\State $\alpha = x^*_y/(b\cdot \sin(angle))$
			\Comment{Computing the scaling $\alpha$ from projection of $x^*_y$ to the minor axis}
			\State $C = \alpha \cdot C_0, d = d_0$
			\State $\kappa_j\gets C, d$
			\State \textbf{return} $\kappa_j$
			
		\end{algorithmic}
	\end{algorithm}
	
	\section*{List of Symbols}\label{sec:los}
	We list the symbols used in the manuscript in the following tables with descriptions.
	\begin{table}[H]
		\centering
		\caption{Description of Symbols}
		\begin{tabular}{@{}cl@{}}
			\hline
			Symbol &Description \\
			\hline
			$n$& Number of mobile manipulator robots (MMRs)\\
			$\{\boldsymbol{w}\}$ & The world fixed reference frame\\
			$\{\boldsymbol{o}\}$ & Object coordinate frame attached to the object \\
			& center of mass \\
			$\{\boldsymbol{b}_i\}$ & Body coordinates attached to the center of $i$-th\\
			&  mobile base\\
			$q_{m,i}$ & Pose of the mobile base of $i$-th MMR\\
			$p_i$ & Position of the mobile base in $\mathbb{R}^2$ of $i$-th MMR \\
			$\phi_i$ & Orientation of the mobile base in $\mathbb{R}$ of $i$-th MMR \\
			$n_i$ & Number of joint of $i$-th MMR \\
			$q_{a,i}$ & Joint displacement of $i$-th MMR \\
			$q_i$ & Combined pose (mobile base) and manipulator\\
			&  joint displacement of $i$-th MMR \\
			$\dot{q}_i$ & Combined pose (mobile base) and manipulator\\
			&  joint of $i$-th MMR \\
			$p_{ee,i}$ & Position of the $i$-th end effector in $\mathbb{R}^3$ \\
			$\phi_{ee,i}$ & Orientation of the $i$-th end effector in $\mathbb{R}^3$ \\
			$v_{i}$ & Linear velocity of the mobile base of $i$-th MMR\\
			$\omega_{i}$ & Angular velocity of the mobile base of $i$-th MMR\\
			$\dot{q}_{a,i}$ & Joint velocity of the manipulator of the $i$-th MMR\\
			$u_{i}$ & Combined velocity of mobile base and the\\
			& manipulator arm of $i$-th MMR\\
			$k$ & Discrete time step\\
			$\underline{q}_{a,i}$ & $i$-th manipulator's joint position lower limit vector\\
			$\overline{q}_{a,i}$ & $i$-th manipulator's joint position upper limit vector\\
			$\underline{u}_{i}$ & The admissible control lower limits of $i$-th MMR\\
			$\overline{u}_{i}$ & The admissible control upper limits $i$-th MMR\\
			$\mathcal{Q}_i$ &  Admissible displacement of $i$-th MMR\\
			$\mathcal{U}_i$ & Admissible control of $i$-th MMR\\
			$\mathcal{F}$ & Multi-MMR formation \\
			$^or_i$ & Grasp pose of $i$-th EE from the object CoM\\
			& measured in $\{\boldsymbol{o}\}$\\
			$p$ &  The position of the object CoM in $\mathbb{R}^3$\\ 
			$o$ & The orientation of the object CoM in $\mathbb{R}^3$\\
			$Q$ & The configuration of $n$ MMRs\\
			$\mathcal{X}$ & The formation configuration\\ 
			$\mathcal{B}(\mathcal{X})$ & The space occupied by the formation \\
			$W$ & A structured and bounded environment having\\
			& both static and dynamic obstacles \\
			$\mathcal{O}$ & The set of static obstacles in $W$ \\
			$\mathcal{O}_{dil}$ & The set of dilated statics obstacles\\
			$W_{free}$ & Static obstacle-free region $W\setminus\mathcal{O}\in\mathbb{R}^2$\\
			$\mathcal{O}_{dyn}$ & The dynamic obstacles in the environment $W$\\
			$p_s$ & The start position \\
			$p_g$ & The goal position \\
			$S$ & Linear piece-wise static obstacle-free shortest path \\
			$S_i$ & Path segment of $S$ \\
			$r_f$ & The radius of the circle enclosing the formation $\mathcal{F}$ \\
			$\mathcal{V}$ & Nodes of the graph \\
			$\mathcal{E}$ & Edges of the graph \\
			$\mathcal{W}$ & Weight of $\mathcal{E}$ of the graph \\ 
			$\mathcal{G}(\mathcal{V}, \mathcal{E}, \mathcal{W})$ & Graph\\
			$w_i$ & Vertices of the path $S$ \\
			$\mathrm{V}_s$ & Set of visible vertices of the statics obstacles\\
			& from $w_i,\ \forall i$\\
			$\mathrm{P}_{cc,i}$ & Static obstacle-free simple polygon around $S_i$ \\
			$\mathcal{P}_{cc}$ & Set of all simple polygon $\mathrm{P}_{cc,i}$ around $S$\\
			$\mathrm{C}$ &  A $2\times2$ symmetric positive definite matrix to\\
			& maps a unit radius circle to an ellipse\\
			$R$ & Rotation matrix that aligns the ellipse axes to\\
			& the world reference frame axes\\
			$\Lambda = diag(a,b)$ & A diagonal scale matrix \\
			$a$, $b$ & The length of the ellipse semi-major and minor axes\\
			$\mathrm{d}$ & Defines the center of the ellipse. \\
			$\kappa(\mathrm{C},\mathrm{d})$ & Ellipse in the ground plane\\
			\hline
		\end{tabular}
		\label{tab:SymbolTab}
	\end{table}
	
	\begin{table}[H]
		\centering
		\begin{tabular}{@{}cl@{}}
			\hline
			Symbol &Description \\
			\hline
			
			$\mathcal{V}_{cc}$ & Set of concave vertices of a simple polygon\\
			$x^*$ & Nearest concave vertices to $\mathrm{d}$ \\
			$H$ & Hyperplane \\
			$\mathcal{P}_{S}$ &  Set of convex polygon around $S$\\
			$p_r(c_t)$ & Time-normalized smooth trajectory guess \\
			$c_t$ & Normalized time parameter  $\in [0,1]$\\
			$N_h$ & Planning horizon segment\\
			$T_h$ & Planning horizon time\\
			$T_e$ & Execution time\\
			$T_c$ & Discretization time-step\\
			$\mathbf{W_u}$ & Diagonal weight-age matrix for control effort\\
			$\mathbf{W_e}$ & Diagonal weight-age matrix for the  trajectory error\\
			${e^k}$ & Tracking error\\
			$\lambda$ & The index of the nearest reference path segment\\
			$J( \mathcal{X}_k,u_k)$ & The Running cost \\
			$J_{N_h}$ & The terminal cost \\
			$v_{op}$ & Operational velocity of the formation\\
			$\mathscr{V}(\mathcal{X})$ & Set of vertices of the bounding polygons of the object and the $n$ MMRs \\
			$d_{safe}$ & The safety distance for static obstacle avoidance \\
			$d_{safe, dyn}$ & The safety distance for dynamic obstacle avoidance \\
			\hline
		\end{tabular}
	\end{table}
	
	\vspace{30pt}


\end{document}